\documentclass[conference]{IEEEtran}
\IEEEoverridecommandlockouts
\usepackage{cite}
\usepackage{amsmath,amssymb,amsfonts}
\usepackage{algorithmic}
\usepackage{graphicx}
\usepackage{textcomp}
\usepackage{xcolor}
\usepackage{times}
\usepackage{epsfig}
\usepackage{graphicx}
\usepackage{amsmath}
\usepackage{amssymb}
\usepackage{multirow}
\usepackage{hhline}
\usepackage{array}
\newcolumntype{P}[1]{>{\centering\arraybackslash}p{#1}}
\newcolumntype{M}[1]{>{\centering\arraybackslash}m{#1}}
\newcolumntype{L}[1]{>{\raggedright\let\newline\\\arraybackslash\hspace{0pt}}m{#1}}
\newcolumntype{C}[1]{>{\centering\let\newline\\\arraybackslash\hspace{0pt}}m{#1}}
\newcolumntype{R}[1]{>{\raggedleft\let\newline\\\arraybackslash\hspace{0pt}}m{#1}}
\usepackage{tablefootnote}
\usepackage{color}
\usepackage{subfig}
\usepackage[linesnumbered,boxed]{algorithm2e}
\usepackage{wrapfig}
\usepackage{tikz}

\usepackage{comment}
\usepackage{flushend}

\def\BibTeX{{\rm B\kern-.05em{\sc i\kern-.025em b}\kern-.08em
    T\kern-.1667em\lower.7ex\hbox{E}\kern-.125emX}}
\begin{document}

\title{Understanding the Impact of Label Granularity on CNN-based Image Classification \\
}

\author{\IEEEauthorblockN{ Zhuo Chen$^1$, Ruizhou Ding$^1$, Ting-Wu Chin, Diana Marculescu}
\IEEEauthorblockA{\textit{Electrical and Computer Engineering} \\
\textit{Carnegie Mellon University}\\
Pittsburgh, USA \\
\{tonychen, rding, tingwuc, dianam\}@cmu.edu}
}

\maketitle
\thispagestyle{plain}
\pagestyle{plain}

\begin{abstract}
In recent years, supervised learning using Convolutional Neural Networks (CNNs) has achieved great success in image classification tasks, and large scale labeled datasets have contributed significantly to this achievement. However, the definition of a \textit{label} is often application dependent. For example, an image of a cat can be labeled as ``cat" or perhaps more specifically ``Persian cat." We refer to this as \textit{label granularity}. In this paper, we conduct extensive experiments using various datasets to demonstrate and analyze how and why training based on fine-grain labeling, such as ``Persian cat" can improve CNN accuracy on classifying coarse-grain classes, in this case ``cat." 

The experimental results show that training CNNs with fine-grain labels improves both network's optimization and generalization capabilities, as intuitively it encourages the network to learn more features, and hence increases classification accuracy on coarse-grain classes under all datasets considered. Moreover, fine-grain labels enhance data efficiency in CNN training. For example, a CNN trained with fine-grain labels and only 40\% of the total training data can achieve higher accuracy than a CNN trained with the full training dataset and coarse-grain labels. These results point to two possible applications of this work: (i) with sufficient human resources, one can improve CNN performance by re-labeling the dataset with fine-grain labels, and (ii) with limited human resources, to improve CNN performance, rather than collecting more training data, one may instead use fine-grain labels for the dataset. We also observe that the improvement brought by fine-grain labeling varies from dataset to dataset, therefore we further propose a metric called \textit{Average Confusion Ratio} to characterize the effectiveness of fine-grain labeling, and show its use through extensive experimentation. Code is available at https://github.com/cmu-enyac/Label-Granularity.

\end{abstract}


\stepcounter{footnote}\footnotetext{Authors contributed equally}

\section{Introduction}

\begin{figure}[t]
    \centering
    \includegraphics[width=\linewidth]{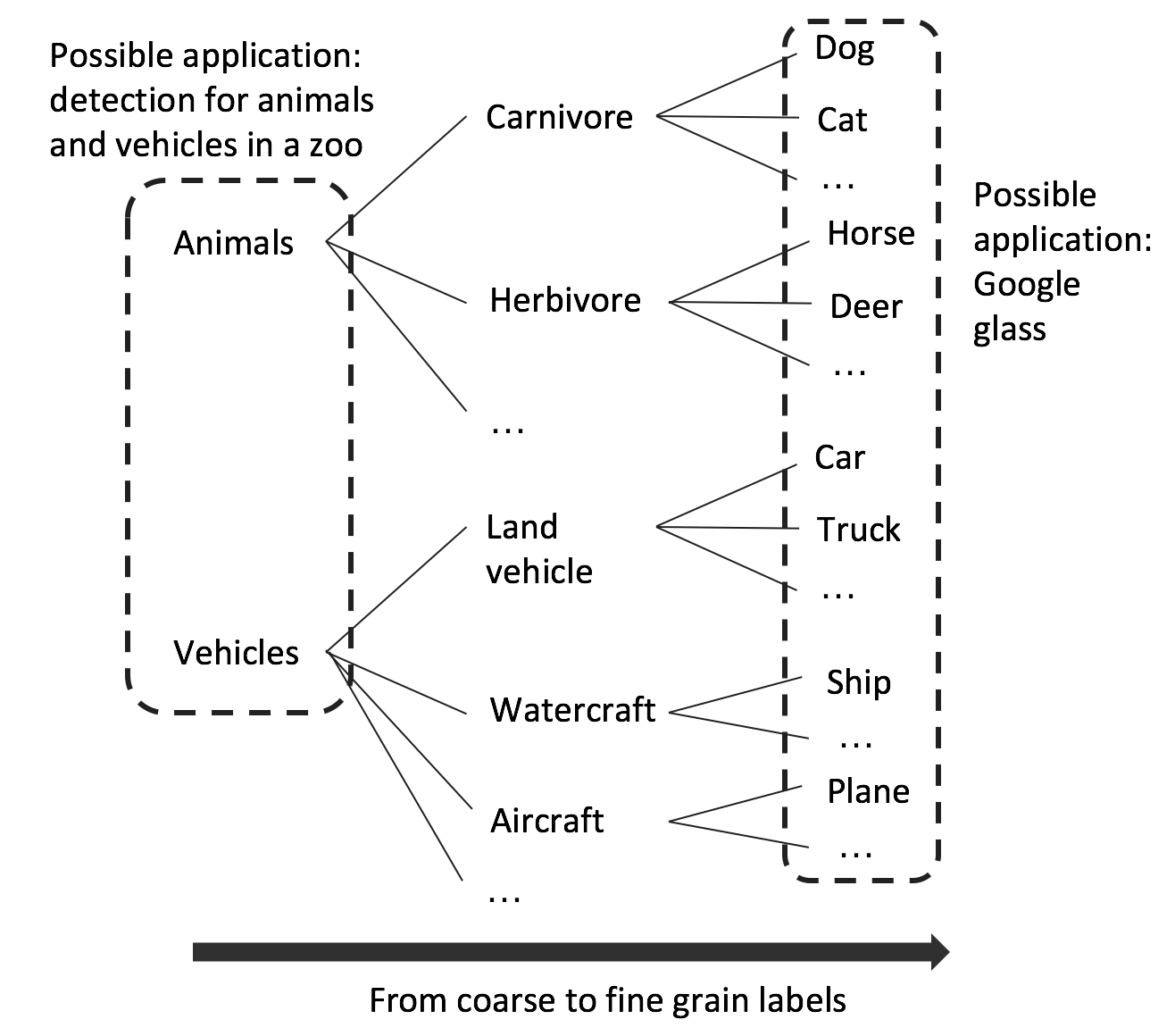}
    \caption{An example of label granularity (label hierarchy). For example, an image of a dog can be labeled ``animal" or ``carnivore" or ``dog", and it is the target application that determines which label to use. This paper explores whether one should use the targeted coarse-grain labels or finer-grain labels for CNN training.}
    \label{fig:im_hier}

\vspace{-15pt}
\end{figure}

We have witnessed tremendous improvement in image classification tasks in recent years thanks to the use of supervised learning combined with the powerful model of Convolutional Neural Networks (CNNs)~\cite{lecun2015deep}. At the same time, the use of large-scale labeled datasets is one of the key elements that has led to this breakthrough \cite{deng2009imagenet} \cite{krizhevsky2009learning}. However, the definition of label varies from application to application, and there is hardly a universal definition of what a ``correct" label is for an image. One such example is how detailed the label should be, \textit{i.e.}, label granularity (or label hierarchy), as illustrated in Figure \ref{fig:im_hier}. For example, in the case of animal image classification, it may be sufficient to label all images of carnivores as ``carnivore", while in an application of carnivore classification, we may label different images as ``dog", ``cat", \textit{etc.}, which are fine-grain labels of the coarse-grain label ``carnivore". Therefore, it is equally correct to label the image of a dog as ``carnivore" or ``dog", yet deciding on which label of the two should be used depends on the task. We denote a fine-grain (coarse-grain) class as a class of images that are labeled with the respective fine-grain (coarse-grain) label, and fine-grain (coarse-grain) training as the training process of CNNs using fine-grain (coarse-grain) labels. If the task at hand is classifying coarse-grain classes, \textit{e.g.}, ``carnivore" vs. ``herbivore", the following question arises: should we directly train and test a CNN using coarse-grain labels as it has usually been done, or would it be beneficial if we trained a CNN with fine-grain labels, \textit{e.g.}, ``dog", ``cat", ``horse", ``deer", \textit{etc.} and map them back to coarse-grain labels during testing phase? The first approach is a method commonly used in image classification tasks \cite{simonyan2014very}, however, in our experiments, we find that training CNNs with fine-grain labels can achieve higher accuracy than using coarse-grain labels in most of the datasets considered. Table \ref{table:prior_new} shows both training and testing accuracy of coarse-grain classification using either coarse-grain or fine-grain labeling. We can see that fine-grain labeling helps improve both training accuracy (network optimization), and testing accuracy (network generalization) across representative image datasets: CIFAR-10~\cite{krizhevsky2009learning}, CIFAR-100~\cite{krizhevsky2009learning}, and ImageNet~\cite{deng2009imagenet}. Moreover, helped by fine-grain labeling, the training process converges faster and requires less amount of training data to achieve the same level of testing accuracy, \textit{i.e.}, becomes more data efficient. More specifically, for the CIFAR-10 dataset and two ImageNet subsets, a CNN trained with fine-grain labels and only 40\% of the total training data can achieve even higher accuracy than a CNN trained with full training dataset but coarse-grain labels. 

In this paper, we design and conduct extensive experiments on various datasets to investigate this interesting phenomenon, and analyze and shed some light on how and why fine-grain label helps enhance coarse-grain image classification. We further propose a metric called \textit{Average Confusion Ratio} (ACR) to characterize the accuracy improvement of fine-grain training, and verify its effectiveness through extensive experiments under different datasets. Our results show two potential practical use of this work: (i) when human resources are abundant, we can increase CNN accuracy by re-labeling the dataset with fine-grain labels and train the CNN using these new labels, and (ii) when human resources are limited and training data is hard to obtain, rather than relying on collecting more training data to improve CNN accuracy, we may instead re-label the dataset with fine-grain labels.

\begin{table}[t]
\begin{center}
\caption{Training and testing accuracy of five datasets when trained with fine-grain labeling (bottom row for each dataset) vs. coarse-grain labeling (top row for each dataset), and tested on coarse-grain labels.}
\vspace{3pt}
\label{table:prior_new}
\begin{tabular}{|C{1.6cm}|C{1.cm}|C{0.9cm}|C{1.3cm}|C{1.3cm}|}
\hline
Dataset & \# of training classes & \# of testing classes & Training accuracy (\%) & Testing accuracy (\%) \\
\hline
\multirow{2}{*}{CIFAR-10} & 2 & 2 & 99.9 & 98.42  \\
\hhline{~----}
& 10 & 2 & 100.0 & 99.20  \\
\hline
\multirow{2}{*}{CIFAR-100} & 20 & 20 & 100.0 & 85.04  \\
\hhline{~----}
& 100 & 20 & 100.0 & 85.05  \\
\hline
\multirow{2}{*}{\parbox{1.5cm}{\centering CIFAR-100 animals}} & 10 & 10 & 100.0 & 81.42  \\
\hhline{~----}
& 50 & 10 & 100.0 & 83.44   \\
\hline
\multirow{2}{*}{\parbox{1.5cm}{\centering ImageNet dog vs. cat}} & 2 & 2 & 94.1 & 92.68  \\
\hhline{~----}
& 10 & 2 & 95.3 & 94.67  \\
\hline
\multirow{2}{*}{\parbox{1.6cm}{\centering ImageNet fruit vs. vege.}} & 2 & 2 & 91.8 & 89.65  \\
\hhline{~----}
& 17 & 2 & 95.4 & 93.15  \\
\hline
\end{tabular}
\end{center}

\vspace{-15pt}
\end{table}

\section{Related Work}

To the best of our knowledge, this is the first work to analyze the use of finer-gain labeling for improving accuracy and training data efficiency for CNN-based image classification tasks. Though there has been significant prior work looking into the hierarchy of classes/categories~\cite{sun2001hierarchical,dekel2004large,bi2012hierarchical,song2014dataless,oh2017top,wang2017local,zhao2017hierarchical,wang2017supervised,cesa2006incremental,deng2012hedging,hoyoux2016can,cerri2015hierarchical}, our work has a distinct objective compared to prior art. Some of the prior work~\cite{cesa2006incremental,dekel2004large,zhao2017hierarchical} aim to utilize the hierarchical label information to improve classification accuracy for the finest categories. On the theory side, Dekel \textit{et al.}~\cite{dekel2004large} propose a learning framework using large margin kernel methods and Bayesian analysis to deal with the classification problem with hierarchical label structures. Cesa-Bianchi \textit{et al.}~\cite{cesa2006incremental} propose a new loss function and use Support Vector Machine (SVM) as well as a probabilistic data model so that higher accuracy can be achieved with exponentially fast convergence speed. From a more practical viewpoint, Zhao \textit{et al.}~\cite{zhao2017hierarchical} leverage hierarchical information of the class structure and select different feature subsets for super-classes. Other work~\cite{deng2012hedging}\cite{bi2012hierarchical}\cite{wang2017local} aim to predict either coarse- or fine-grain labels conditioning on the confidence level. Deng \textit{et al.}~\cite{deng2012hedging} optimize the trade-off between specificity (how fine-grain the predicted label is) and accuracy, while Bi \textit{et al.}~\cite{bi2012hierarchical} develop a Bayes-optimal classifier to minimize the Bayesian risk. More recently, Wang \textit{et al.} propose to stop the prediction process for a coarse-grain label so as to avoid an incorrect prediction. In addition, other prior work~\cite{song2014dataless}\cite{hoyoux2016can}\cite{oh2017top} focuses on the understanding of hierarchical labels. For example, Song \textit{et al.}~\cite{song2014dataless} study dataless hierarchical text classification with unsupervised methods. Hoyoux \textit{et al.}~\cite{hoyoux2016can} show some counter-examples where using hierarchical methods degrades the accuracy, and explore the reasons for such results. Oh~\cite{oh2017top} studies the combination of hierarchical classification and top-\textit{k} accuracy.

However, all these studies aim to increase the classification accuracy of fine-grain classes. Instead, we focus on the case of coarse-grain classes being the target of classification task, and we explore whether directly training with finer-grain labels can achieve higher classification accuracy on coarse-grain classes than training with coarse-grain labels. 

A work close to ours is done by Mo \textit{et al.}~\cite{mo2016learning}, who propose active over-labeling to generate finer-grain labels than the target coarse-grain labels, and demonstrate that fine-grained label data can improve precision of a classifier for the coarse-grained concept. Similar ideas were also studied by other prior work~\cite{hoffmann2001using,luo2008can,fradkin2008clustering,ahmed2012estimating,ristin2015categories}. However, none of them explores deep learning models which are very different from conventional machine learning models, \textit{e.g.}, Support Vector Machine (SVM), logistic regression, etc. Fradkin~\cite{fradkin2008clustering} performs experiments on linear and non-linear SVM, and finds that fine-grain training can improve accuracy for linear SVM since fine-grain labeling can learn a piece-wise linear decision boundary that better approximate the true non-linear boundary. However, fine-grain training does not help non-linear (RBF-kernel) SVM due to their inherent non-linearity. CNNs are highly non-linear models and relatively more difficult to optimize~\cite{glorot2010understanding}. No results of CNNs has yet been shown on this topic. 

The remainder of this paper is organized as follows. Section \ref{sec:label_granularity} demonstrates in detail the effects of fine-grain labeling improving CNN-based image classification and its capability of enhancing training data efficiency. Section \ref{sec:explanation} analyzes why fine-grain labeling helps in terms of both network \textit{optimization} and \textit{generalization} via extensive experiments on various datasets. In Section \ref{sec:acr}, we propose a metric called \textit{Average Confusion Ratio} to characterize the accuracy gain of fine-grain training, and verify its effectiveness under different datasets. In Section \ref{sec:discuss}, we further discuss how (i) customized coarse-grain classes for diverse applications, (ii) noisy fine-grain labels obtained via automatic clustering methods, and (iii) the number of coarse-grain classes may impact effectiveness of fine-grain training. We conclude our work in Section \ref{sec:conclusion}.
\section{Label Granularity and Training Data} \label{sec:label_granularity}

In this section, we demonstrate the effects of fine-grain labels on improving image classification accuracy and further show its capability of enhancing training efficiency.

We define $A_{FC}^{train}$ and $A_{FC}^{test}$ as the training and testing accuracy of a CNN trained on fine-grain labels and evaluated on coarse-grain labels, respectively. In detail, we first train a network with fine-grain labels and output the predicted fine-grain labels of all input images. Then we map the predicted fine-grain labels to their respective coarse-grain labels via the predefined mapping as shown in table \ref{table:class_setting}. Finally, the accuracy is computed by comparing the predicted coarse-grain labels with the ground-truth labels. Similarly, we define $A_{CC}^{train}$ and $A_{CC}^{test}$ as the training and testing accuracy of a CNN trained on coarse-grain labels and evaluated on the same labels. To do this, we directly train a network with coarse-grain labels and compute accuracy by comparing the predicted labels, which are already coarse-grain labels, of the input images with their ground-truth labels.

We design and conduct experiments on well-known image classification datasets:  CIFAR-10~\cite{krizhevsky2009learning}, CIFAR-100~\cite{krizhevsky2009learning} and ImageNet~\cite{deng2009imagenet}, and we list their coarse- and fine-grain classes in Table \ref{table:class_setting}. CIFAR-10 dataset is a great fit for applications similar to the one shown in Figure \ref{fig:im_hier}, \textit{i.e.}, classifying whether an image contains an animal or a vehicle. CIFAR-10 has six animals: ``bird", ``cat", ``deer", ``dog", ``frog", ``horse", and four vehicles: ``plane",``car", ``ship", ``truck". CIFAR-100 provides 20 coarse-grain classes and five fine-grain classes per coarse-grain class, resulting in 100 fine-grain classes in total. We also select all ten animal coarse-grain classes from CIFAR-100 to form another dataset serving applications like animal classification, and we call this dataset: CIFAR-100 animals. ImageNet dataset is collected and organized according to the WordNet hierarchy~\cite{deng2009imagenet, miller1995wordnet} and therefore it naturally follows the coarse-to-fine-grain label hierarchy. We use subsets of ImageNet dataset to better visualize and demonstrate the benefits of training CNN with fine-grain labels. The first ImageNet subset task is to classify dog vs. cat, with a total of ten fine-grain classes of random breeds of dogs and cats. The second task is classifying fruit vs. vegetable with a total of 17 fine-grain classes.

\begin{table}[ht]
\begin{center}
\caption{Coarse-grain and fine-grain classes of five datasets.}
\vspace{3pt}
\label{table:class_setting}
\begin{tabular}{|C{1.6cm}|C{2cm}|C{3.8cm}|}
\hline
Dataset & Coarse-grain classes & Fine-grain classes \\
\hline
\multirow{2}{*}{CIFAR-10} & animal & bird, cat, deer, dog, frog, horse \\
\hhline{~--}
& vehicle & plane, car, ship, truck \\
\hline
\multirow{30}{*}{CIFAR-100} & aquatic mammals* & beaver, dolphin, otter, seal, whale \\
\hhline{~--}
& fish* & aquarium fish, flatfish, ray, shark, trout \\
\hhline{~--}
& flowers & orchid, poppy, rose, sunflower, tulip \\
\hhline{~--}
& food containers & bottle, bowl, can, cup, plate \\
\hhline{~--}
& fruit and vegetables & apple, mushroom, orange, pear, sweet pepper \\
\hhline{~--}
& household electrical devices & clock, keyboard, lamp, telephone, television \\
\hhline{~--}
& household furniture & bed, chair, couch, table, wardrobe \\
\hhline{~--}
& insects* & bee, beetle, butterfly, caterpillar, cockroach \\
\hhline{~--}
& large carnivores* & bear, leopard, lion, tiger, wolf \\
\hhline{~--}
& large man-made outdoor things & bridge, castle, house, road, skyscraper \\
\hhline{~--}
& large natural outdoor scenes & cloud, forest, mountain, plain, sea \\
\hhline{~--}
& large omnivores and herbivores* & camel, cattle, chimpanzee, elephant, kangaroo \\
\hhline{~--}
& medium mammals* & fox, porcupine, possum, raccoon, skunk \\
\hhline{~--}
& non-insect invertebrates* & crab, lobster, snail, spider, worm \\
\hhline{~--}
& people* & baby, boy, girl, man, woman \\
\hhline{~--}
& reptiles* & crocodile, dinosaur, lizard, snake, turtle \\
\hhline{~--}
& small mammals* & hamster, mouse, rabbit, shrew, squirrel \\
\hhline{~--}
& trees & maple tree, oak tree, palm tree, pine tree, willow tree \\
\hhline{~--}
& vehicles 1 & bicycle, bus, motorcycle, pickup truck, train \\
\hhline{~--}
& vehicles 2 & lawn mower, rocket, streetcar, tank, tractor \\
\hline
\parbox{1.5cm}{\centering CIFAR-100 animals} & (10 coarse-grain classes above marked with *) & (50 corresponding fine-grain classes) \\
\hline
\multirow{3}{*}{\parbox{1.5cm}{\centering ImageNet dog vs. cat}} & dog & basset, chihuahua, maltese, papillon, pekinese,  \\
\hhline{~--}
& cat &  tabby, tiger cat, Persian, Siamese, Egyptian \\
\hline
\multirow{5}{*}{\parbox{1.5cm}{\centering ImageNet fruit vs. vege}} & fruit & strawberry, orange, lemon, fig, pineapple, banana, jackfruit, custard apple  \\
\hhline{~--}
& vege &  head cabbage, broccoli, cauliflower, zucchini, butternut squash, cucumber, artichoke, pepper, mushroom \\
\hline
\end{tabular}
\end{center}

\vspace{-15pt}
\end{table}

\begin{table}[ht]
\begin{center}
\caption{Configuration of CNNs used in the experiments.}
\vspace{3pt}
\label{table:network}
\begin{tabular}{|C{2cm}|C{1.6cm}|C{1.6cm}|C{1.6cm}|}
\hline
Dataset & \# of layers & \# of filters in widest layer & \# of parameters \\
\hline
CIFAR-10 & 18 & 512 & 11.1M \\
\hline
CIFAR-100 CIFAR-100 animals & 26 & 640 & 36.5M \\
\hline
ImageNet subsets & 18 & 128 & 0.7M \\
\hline
\end{tabular}
\vspace{-6pt}
\end{center}
\end{table}

\begin{figure}[ht]
    \centering
    \subfloat[][CIFAR-10]{\includegraphics[width=.5\linewidth]{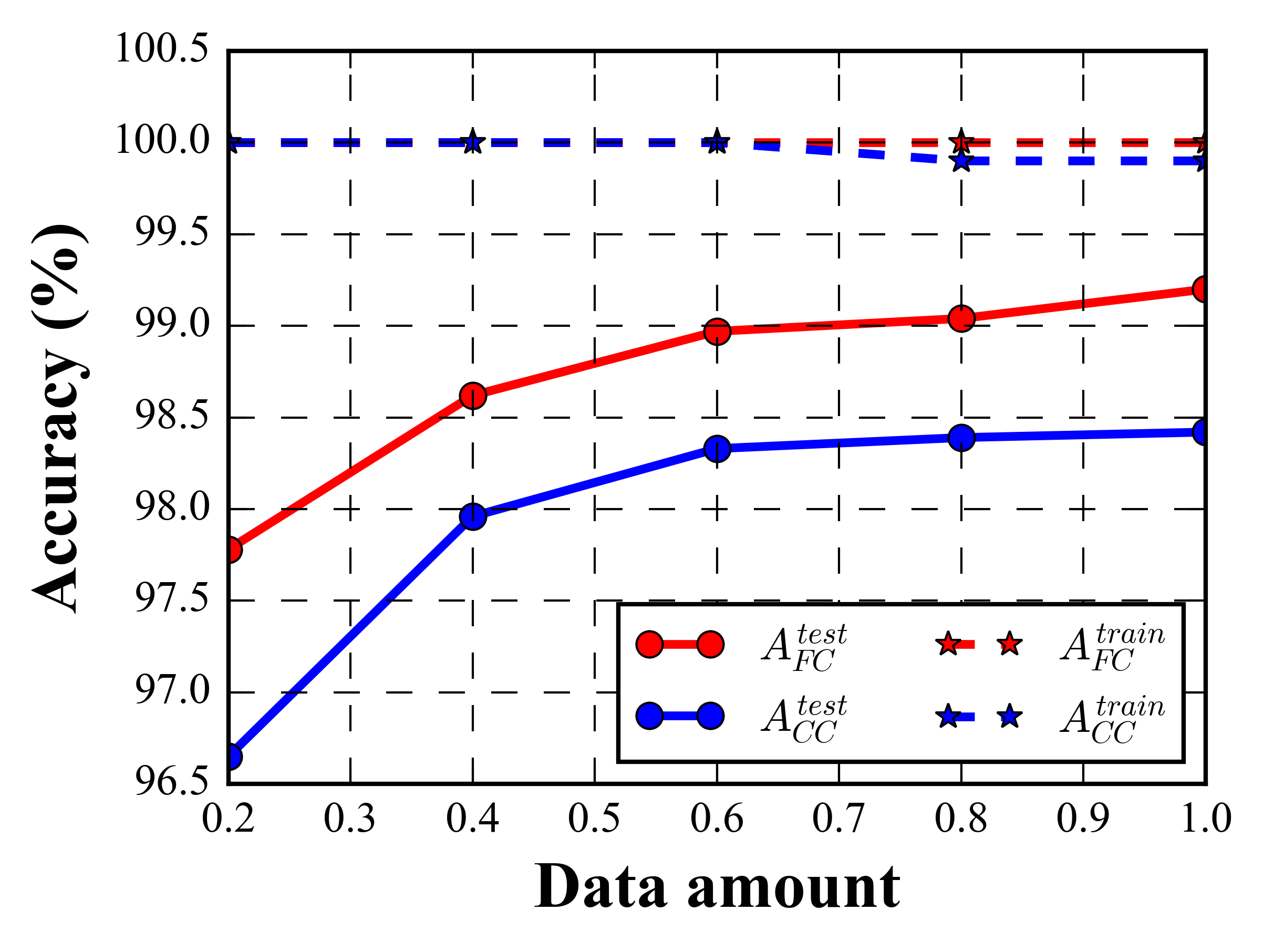}\label{fig:cifar10_datanum}}
    \subfloat[][CIFAR-100]{\includegraphics[width=.5\linewidth]{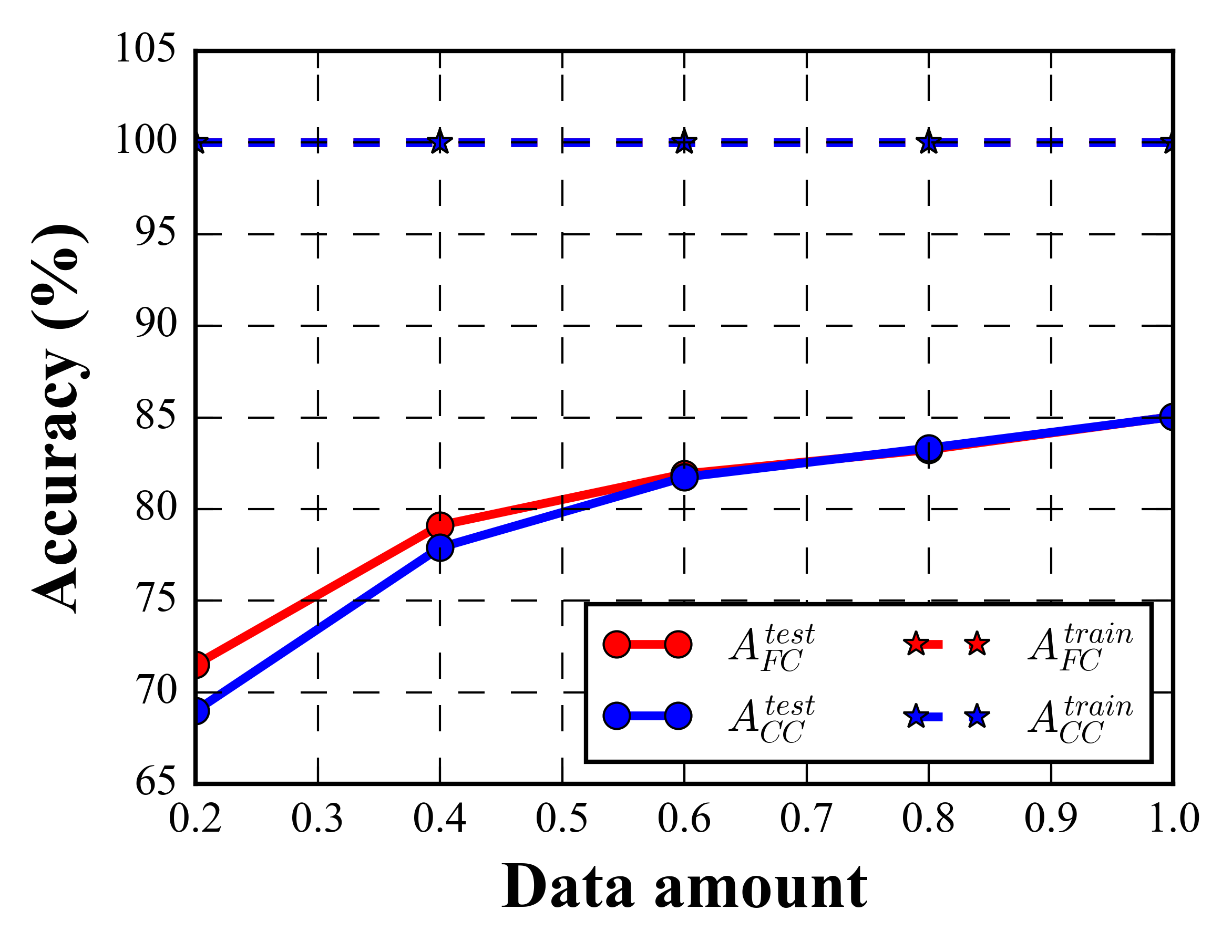}\label{fig:cifar100_datanum}}
    
    \subfloat[][CIFAR-100-animal]{\includegraphics[width=.5\linewidth]{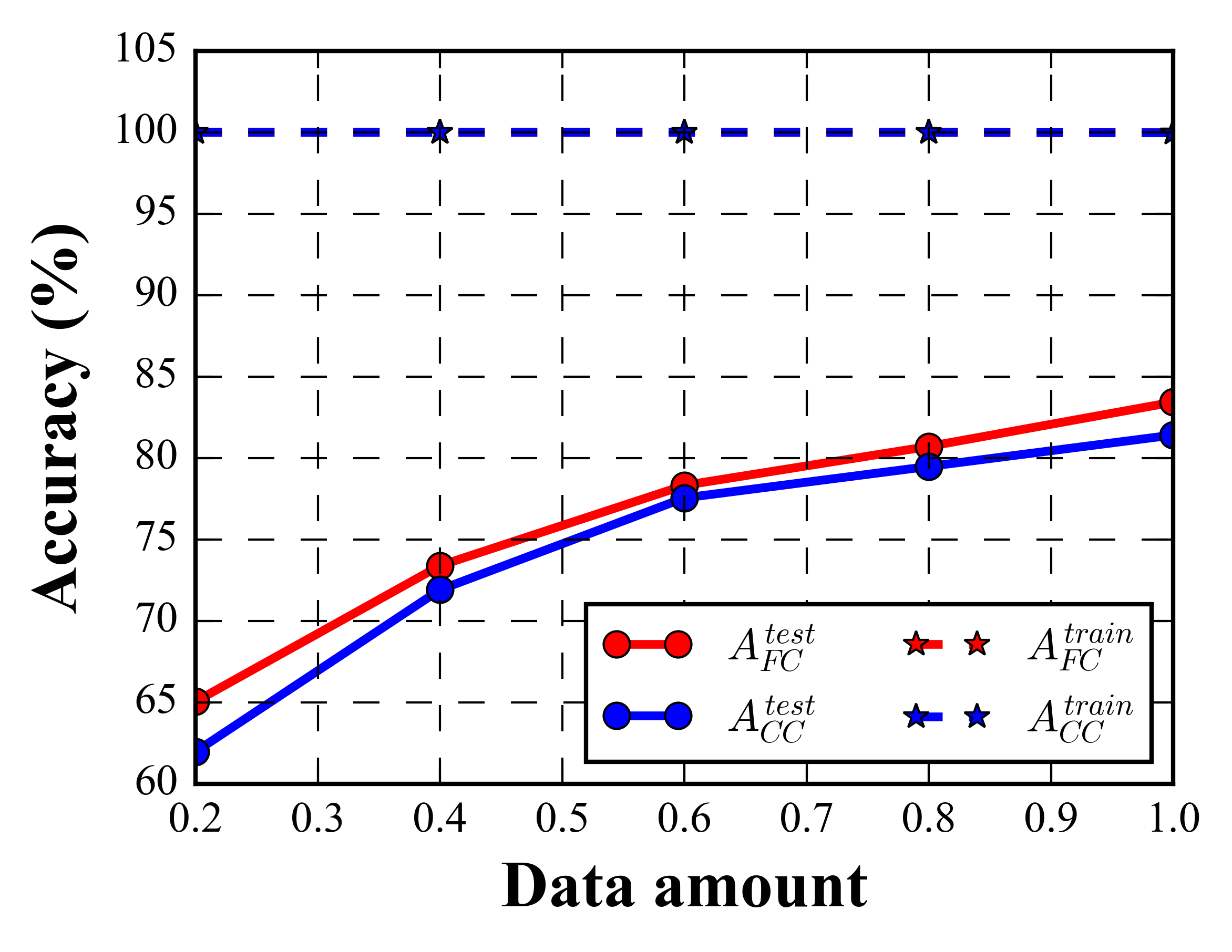}\label{fig:cifar100_animal_datanum}}
    \subfloat[][Dog vs. cat]{\includegraphics[width=.5\linewidth]{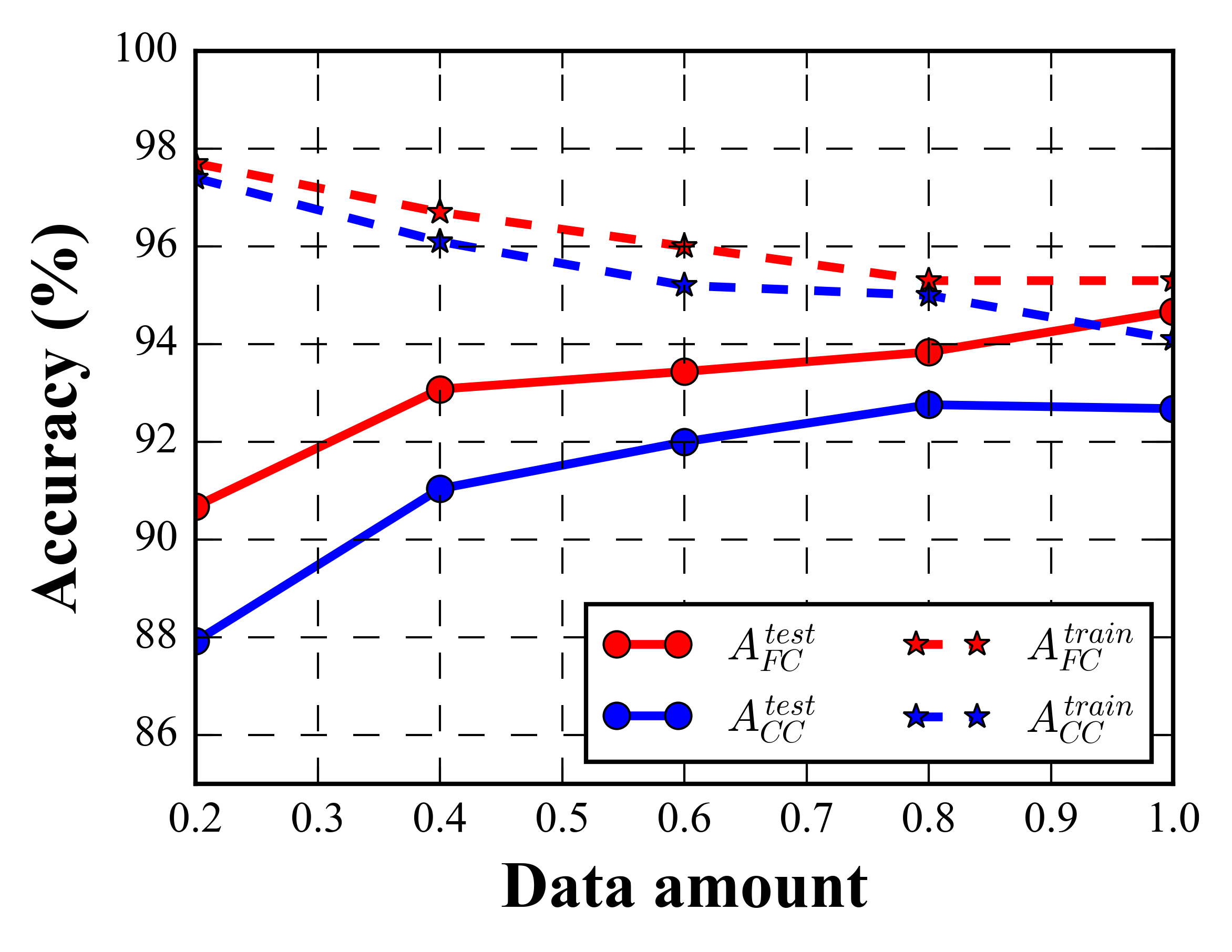}\label{fig:dog_datanum}}
    
    \subfloat[][Fruit vs. vegetable]{\includegraphics[width=.5\linewidth]{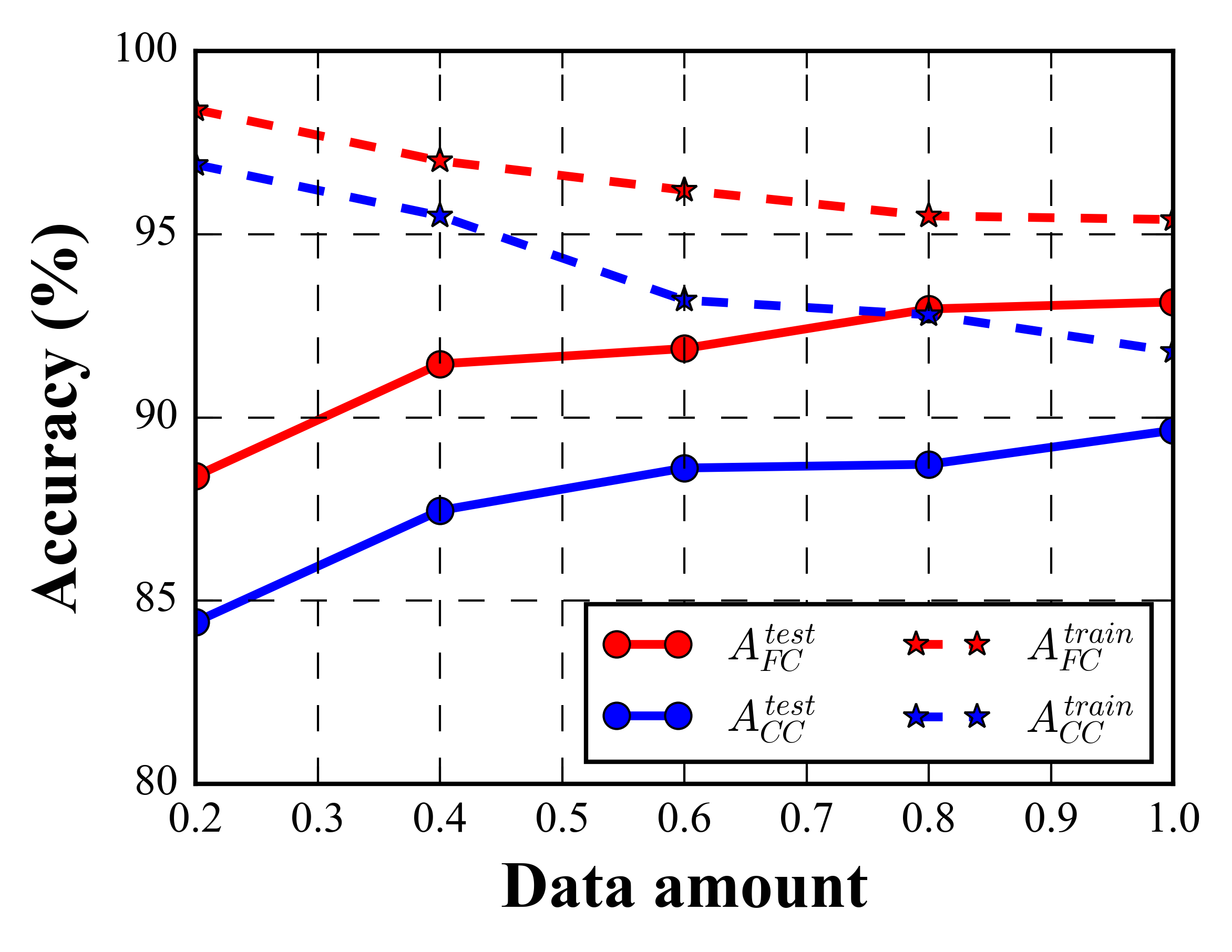}\label{fig:fruit_datanum}}
    \caption{Training (dotted) and testing (solid) accuracy curves with increasing amount of training data. CNNs trained with fine-grain labels are shown in red and those trained with coarse-grain labels are shown in blue. Experiments are conducted using five datasets: (a) CIFAR-10, (b) CIFAR-100, (c) CIFAR-100-animal, and two subsets of ImageNet datasets (d) dog vs. cat, (e) fruit vs. vegetable.  }
    \label{fig:data_amount}
 
 \vspace{-5pt}
\end{figure}

Table~\ref{table:network} shows the network configurations used for different datasets. For CIFAR-10, we use the full pre-activation residual network with 512 filters for the widest layer similar to the one in \cite{he2016identity}. We use wide residual network~\cite{zagoruyko2016wide} for CIFAR-100 dataset, which achieves $81.15\%$ accuracy on 100 classes. We use ``thinner" networks for ImageNet subsets to avoid overfitting because ImageNet subsets have fewer training images (22K images for fruit vs. vegetable, and 13K images for dog vs. cat) than CIFAR-10 (50K images) and CIFAR-100 (50K images). The network configuration for ImageNet subsets is similar to CIFAR-10, but with $75\%$ fewer filters per convolution layer. We use random cropping and random flipping data augmentation \cite{he2016deep} for all datasets. For the training configuration, we use momentum 0.9 and weight decay 5e-4. The learning rate starts at 0.1 for CIFAR-10 and CIFAR-100, and 0.01 for ImageNet subsets, and decays when the loss plateaus. We train CIFAR-10 and CIFAR-100 for 200 epochs, and ImageNet subsets for 225 epochs. The above CNN architectures and training schemes are derived from existing works, which are usually near-optimal via extensive hyper-parameter tuning. To keep the comparison fair, when we train on fine-grain labels, we keep everything unchanged, except for the last layer which will need to output more classes. Note that without fine-tuning on our fine-grain training, we are still able to outperform the hand-tuned models.

Table \ref{table:prior_new} shows the results. The second column gives the number of classes CNN is trained on and the third column shows the number of classes the CNN is tested on. If these two numbers are the same, it means training and testing are both using the same coarse-grain labels. If they are different, it means that CNN is trained first with fine-grain labels and then tested on the coarse-grain labels. We can see that training using fine-grain labels almost always improves testing accuracy compared to training using coarse-grain labels. In the case of CIFAR-100, fine-grain training provides negligible improvement on testing accuracy. We conjecture that this is due to the diminishing return when there are more coarse-grain labels, and we verify this hypothesis in Section~\ref{sec:vary_coarse}. For the CIFAR-10 dataset, although the absolute value of improvement is 0.78\%, considering the original testing accuracy is already near perfect, this further improvement is non-trivial. We also observe that CNN training accuracy gets better when using fine-grain labels. Above results indicate that fine-grain labels help improve both network optimization and generalization, and we will analyze the reasons in Sections \ref{sec:optimization} and \ref{sec:generalization}, respectively. 

We further investigate how fine-grain labels affect training data efficiency. High training data efficiency means that (i) with the same amount of data, CNNs are able to learn and perform better, \textit{i.e.}, achieve higher testing accuracy, and (ii) to achieve the same testing accuracy, CNNs require fewer training data. To this end, we randomly chose $20\%$, $40\%$, $60\%$ or $80\%$ of the entire training dataset to form four new training sets with increasing data amount (same proportion in each class so that the number of images within each class is still balanced), use the full testing dataset for testing, and compare the accuracy of fine-grain and coarse-grain training. Since we find that keeping the same number of epochs for reduced data amounts leads to fewer weight updates, we use proportionally more training epochs for less training data to keep the number of weight updates the same. In other words, when $20\%$ of training data is used, we train for 5X epochs.

The results are depicted in Fig \ref{fig:data_amount}, where we show training and testing accuracy for both fine-grain and coarse-grain training, \textit{i.e.}, $A_{FC}^{train}$, $A_{FC}^{test}$, $A_{CC}^{train}$ and $A_{CC}^{test}$. We observe that training with fine-grain labels almost always improves testing accuracy. Especially in the case of (a), (d) and (e) of Figure \ref{fig:data_amount} (CIFAR-10, ImageNet dog vs. cat, and ImageNet fruit vs. vegetable, respectively), we observe a significant improvement from the use of fine-grain labels: with less than 40\% of the total training data, training with fine-grain labels is able to achieve even higher accuracy than using the full training dataset with coarse-grain labels. For CIFAR-100 animals (c), with only 80\% of the total data amount, training with fine-grain labels is able to achieve comparable accuracy as using coarse-grain labels and full training dataset. Although fine-grain training has negligible improvement on testing accuracy with full dataset in case of CIFAR-100, when using fewer than 40\% of the full dataset, fine-grain training still exhibits a clear advantage. This indicates that when availability of data is limited, having fine-grain labels can be helpful. 

These experimental results show that training with fine-grain labels can help CNNs better utilize the available training data and can almost always improve CNN accuracy. One may think this counter-intuitive, as how could training on more classes require fewer training samples? The reasons are that i) fine-grain labels encourage the CNN to learn more features which helps with generalization, and ii) the test accuracy is eventually evaluated based on coarse-grain labels rather than fine-grain labels. We will discuss in detail on how fine-grain labels improve CNN performance in the following section. These results indicate two potential practical usage of this work: i) if we have sufficient human resources, we can improve CNN performance by re-labeling data with fine-grain labels, and ii) if we have limited human resources, in order to improve CNN performance, it can be more advantageous to re-labeling images with fine-grain labels rather than collecting more data. 

\section{Optimization and Generalization}\label{sec:explanation}

\begin{figure}[t]
    \centering
    \subfloat[][CIFAR-10]{\includegraphics[width=.5\linewidth]{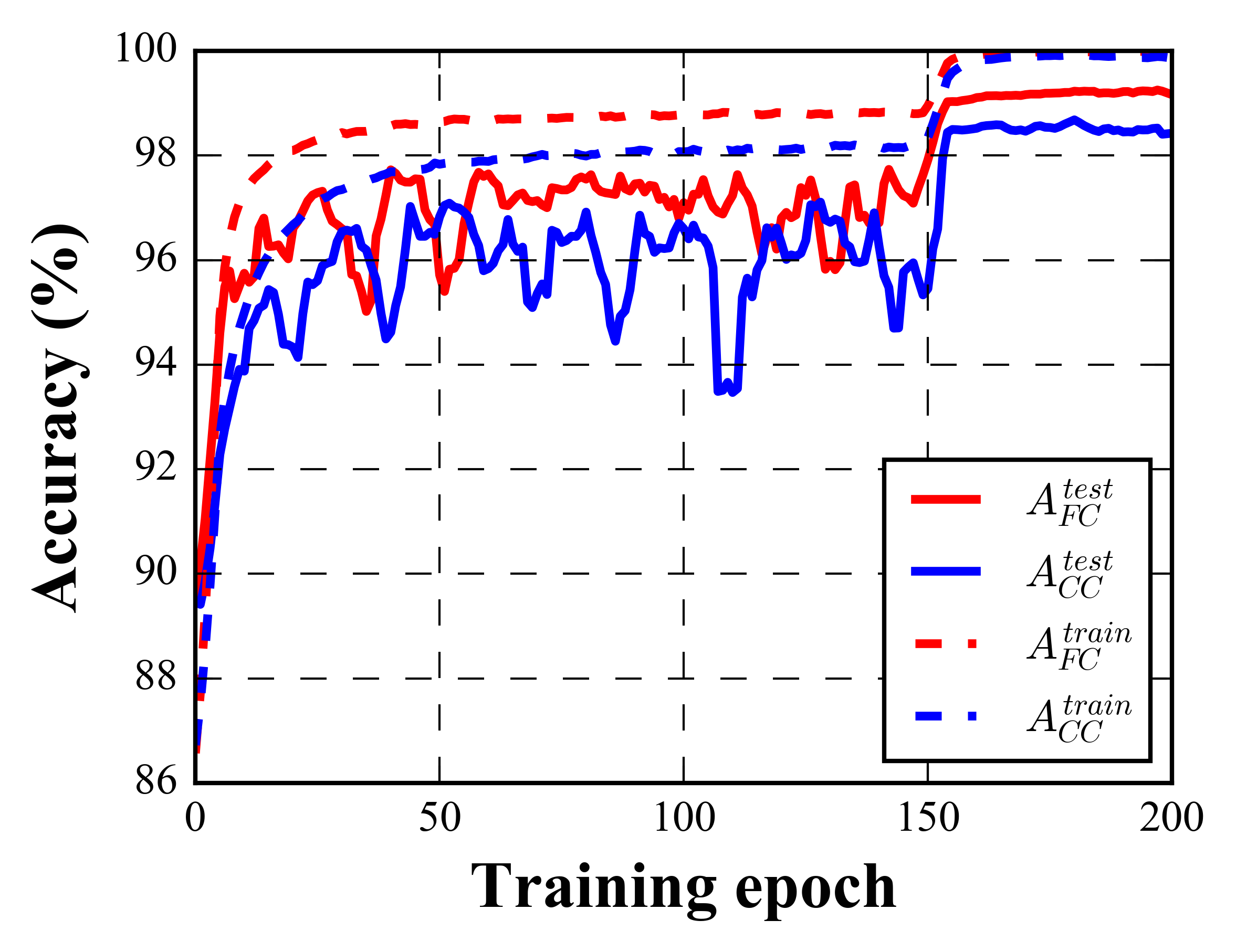}\label{fig:cifar10_curve}}
    \subfloat[][CIFAR-100]{\includegraphics[width=.5\linewidth]{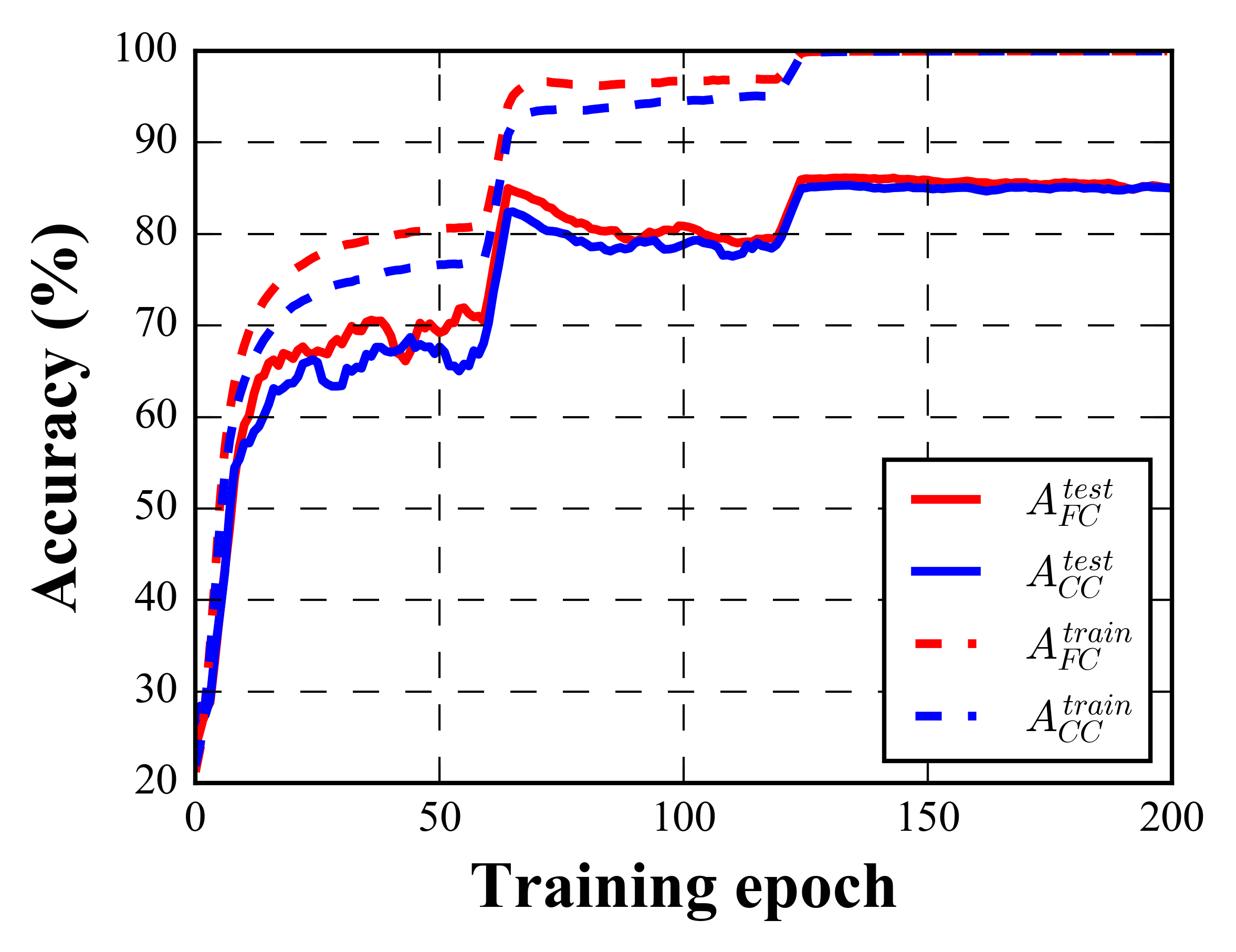}\label{fig:cifar100_curve}}
    
    \subfloat[][CIFAR-100 animals]{\includegraphics[width=.5\linewidth]{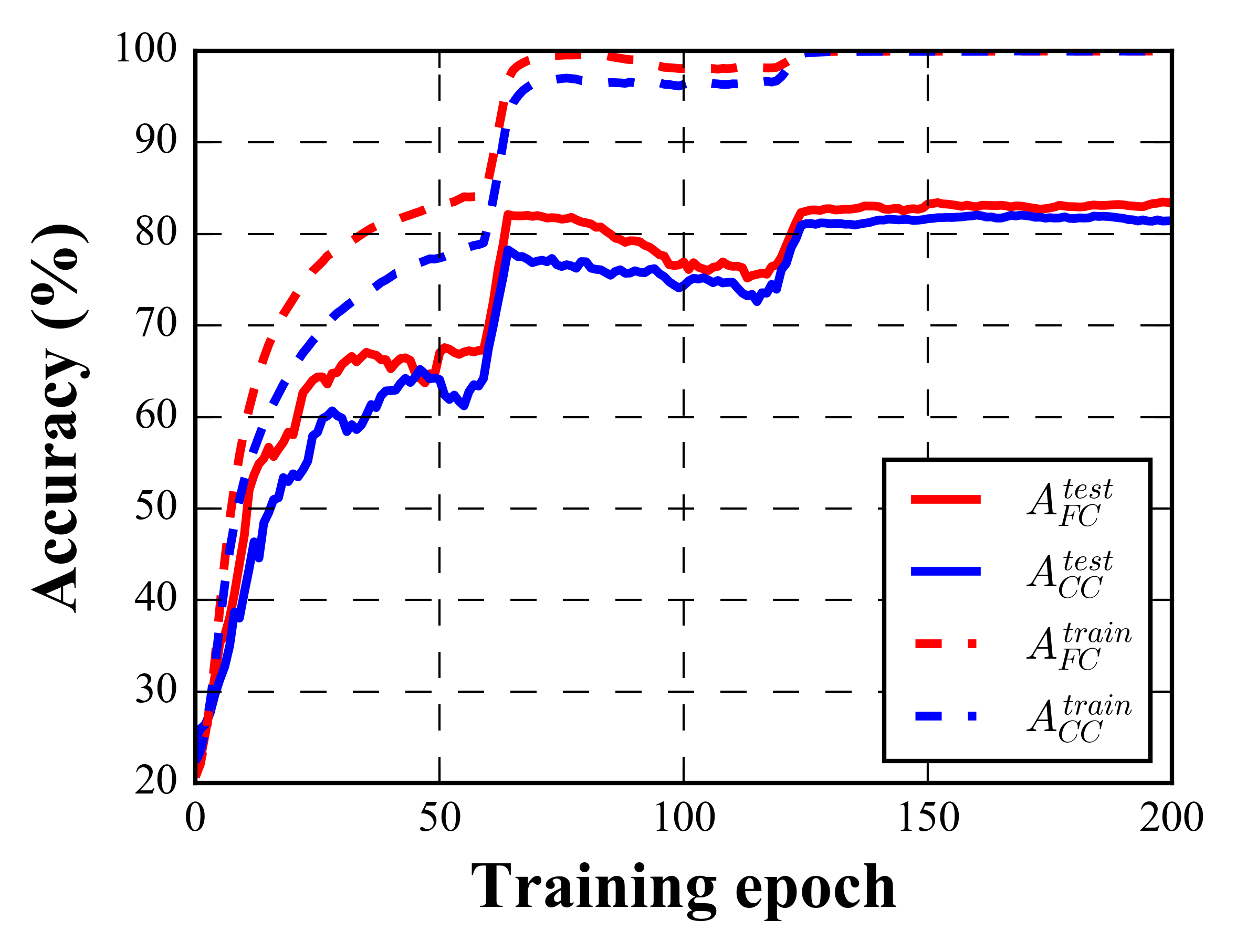}\label{fig:cifar100_animal_curve}}
    \subfloat[][Dog vs. cat]{\includegraphics[width=.5\linewidth]{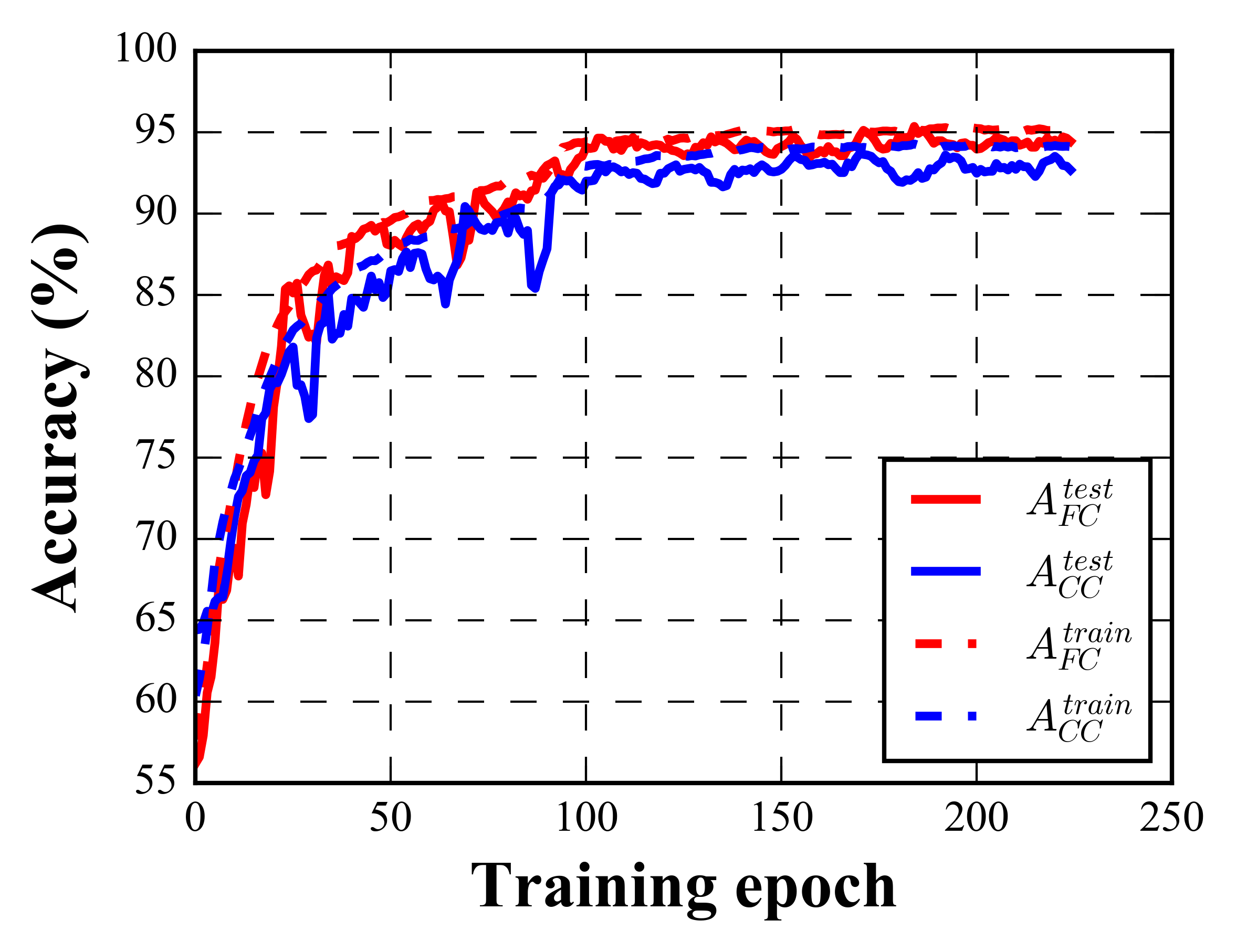}\label{fig:dog_curve}}
    
    \subfloat[][Fruit vs. vegetable]{\includegraphics[width=.5\linewidth]{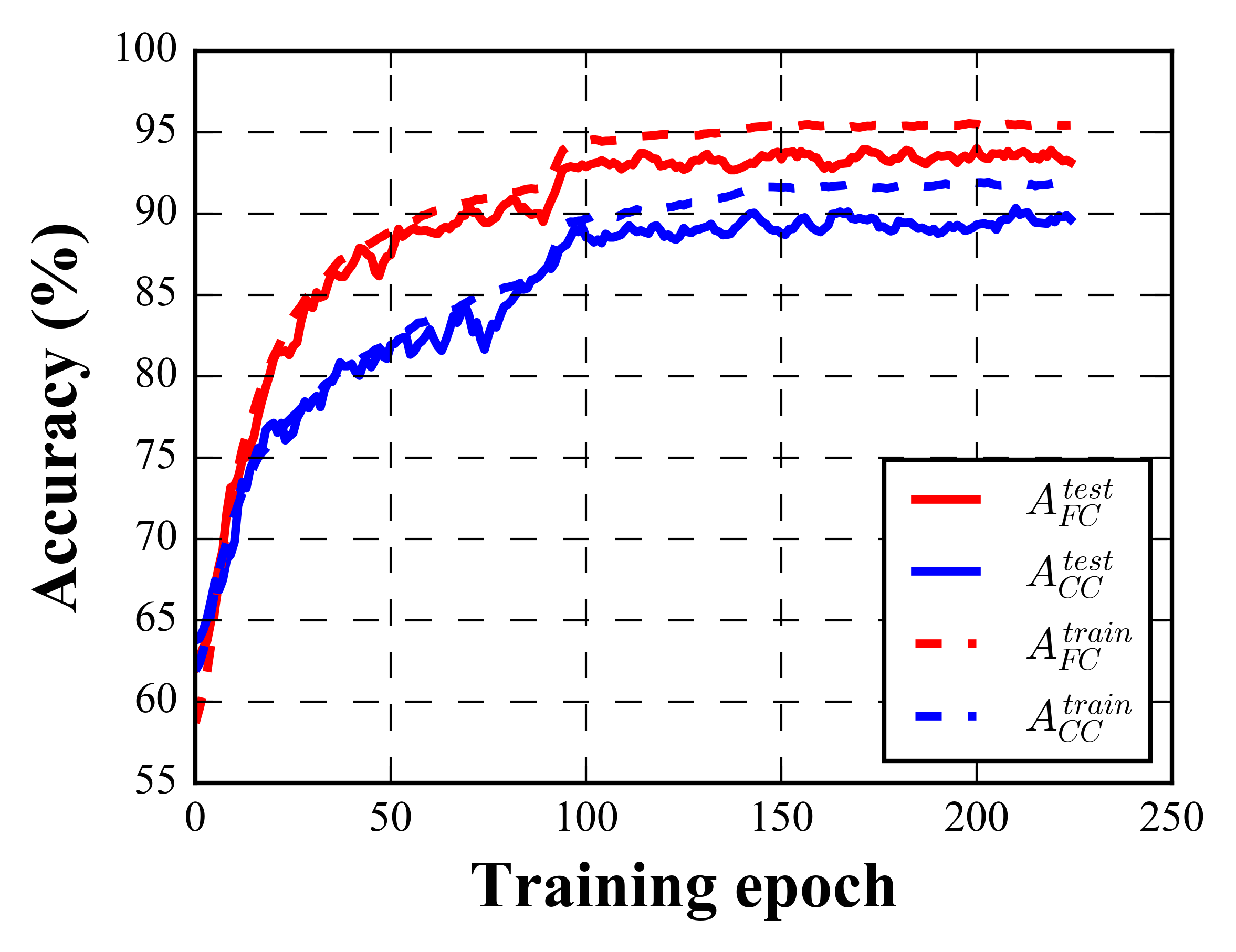}\label{fig:fruit_curve}}
    \caption{Training (dotted) and testing (solid) accuracy curves for five datasets. CNNs trained with fine-grain labels are shown in red and those trained with coarse-grain labels are shown in blue. Experiments are conducted using five datasets: (a) CIFAR-10, (b) CIFAR-100, (c) CIFAR-100 animals, and two subsets of ImageNet datasets (d) dog vs. cat and (e) fruit vs. vegetable. }
    \label{fig:training_curve}
 
 \vspace{-15pt}
\end{figure}

As we have discussed in Section \ref{sec:label_granularity}, training with fine-grain labels can improve not only testing but also training accuracy. This means that fine-grain labels help with both network \textit{optimization} and \textit{generalization}. In this section, we design and conduct extensive experiments on all datasets showing how both optimization and generalization are improved.

\subsection{Optimization}\label{sec:optimization}
In Figure \ref{fig:training_curve}, the dotted curves show the training accuracy of both fine-grain training, $A_{FC}^{train}$, and coarse-grain training, $A_{CC}^{train}$, for all datasets. The training accuracy is evaluated at the end of every epoch during the training phase by using the training dataset. We can see that training with fine-grain labels not only achieves higher training accuracy, but also converges faster as the red curve is always above the blue curve. The accuracy jumps are the results of reduced learning rate and are common phenomena in training neural networks~\cite{he2016deep}.

Prior art investigating fine-grain labels on simple linear classifiers argues that the reason fine-grain labeling helps is the ability to learn piece-wise linear decision boundaries that can better approximate the true non-linear decision boundary~\cite{hoffmann2001using}. That is, fine-grain training can have higher non-linearity compared to coarse-grain training due to increased parameters in the model. However, a further study \cite{fradkin2008clustering} shows that in the case of non-linear classifiers, \textit{e.g.}, RBF-kernel Support Vector Machine (SVM), fine-grain training no longer improves accuracy compared to using coarse-grain labels because the network itself has sufficient non-linearity to learn the non-linear decision boundary without the help of fine-grain labels. In the case of CNNs, we ask the question: is this piece-wise linear nature the reason for better training accuracy for fine-grain labels compared to coarse-grain training?

CNNs are already highly non-linear, so we conjecture that the answer is no. To evaluate this, we insert another fully-connected layer to a coarse-grain trained network right after the global pooling layer, so that compared to the original network, it can also achieve a piece-wise linear boundary on the high-level features. We train the new network end to end from scratch instead of pre-loading and freezing the weights of the preceeding layers, such that it fully utilizes all the degrees of freedoms of the model, possibly achieving higher training accuracy.
We train this new network structure with coarse-grain labels, and compare the results with the baseline network trained with coarse- or fine-grain labels. 
We keep the training scheme for the slightly deeper network the same as that of the baseline network for a fair comparison with coarse- and fine-grain training. 

Table \ref{table:add_linear} shows our results. In the "CNN Arch" column, 'Extra layer' means that we add the fully-connected layer to the baseline CNN as described above. In the "Train Label" column, "F" and "C" indicate fine-grain and coarse-grain labels, respectively. The values in parentheses following each training and testing accuracy value are the improvement/degradation with respect to the training and testing accuracy of a baseline CNN trained with coarse-grain labels, respectively.

We can see that, compared to the baseline CNN trained with coarse-grain labels, adding one extra layer does not bring significant improvement in either optimization or generalization. In certain cases, the testing accuracy is degraded, in CIFAR-100 animals and ImageNet subset dog vs. cat, possibly due to the difficulty in optimizing a larger network.

This means that simply adding non-linearity to coarse-grain training cannot match the training accuracy brought by fine-grain training. That is, the slightly higher non-linearity brought by fine-grain training is not the only reason for achieving higher training accuracy. Rather, it is more likely that fine-grain labels give more hints to the network about which features to learn. This is also supported by the experimental results in Section \ref{sec:random_fine_class}, where we randomly generate fine-grain labels for each coarse-grain class, and find out that fine-grain training does not optimize better than coarse-grain training. 


\begin{table}
\begin{center}
\caption{Experiments on increasing CNN non-linearity and capacity under coarse-grain training. In "CNN Arch": 'Extra layer' means that we add the fully-connected layer to the baseline CNN to increase network non-linearity and capacity as described in Section \ref{sec:optimization}. In "Train Label": "F" and "C" indicate fine-grain and coarse-grain labels, respectively. In the training and testing accuracy columns, the values indicated in the parentheses are the improvement/degradation with respect to the training and testing accuracy of a baseline CNN trained with coarse-grain labels, respectively.}
\vspace{3pt}
\label{table:add_linear}
\begin{tabular}{|C{1.6cm}|C{1.6cm}|C{0.9cm}|C{1.2cm}|C{1.2cm}|}
\hline
Dataset & CNN Arch & Train Label & Training accuracy (\%) & Testing accuracy (\%) \\
\hline
\multirow{2}{*}{CIFAR-10} & Baseline CNN & F & 100.0 (+0.1) & 99.20 (+0.78)  \\
\hhline{~----}
& Extra layer & C & 99.9 (+0.0) & 98.50 (+0.08)  \\
\hline
\multirow{2}{*}{CIFAR-100} & Baseline CNN & F & 100.0 (+0.0) & 85.05 (+0.01)  \\
\hhline{~----}
& Extra layer & C & 100.0 (+0.0) & 86.33 (+1.29)  \\
\hline
\multirow{2}{*}{\parbox{1.5cm}{\centering CIFAR-100 animals}} & Baseline CNN & F & 100.0 (+0.0) & 83.44 (+2.02)  \\
\hhline{~----}
& Extra layer & C & 100.0 (+0.0) & 80.73 (-0.69)  \\
\hline
\multirow{2}{*}{\parbox{1.5cm}{\centering ImageNet dog vs. cat}} & Baseline CNN & F & 95.3 (+1.2) & 94.87 (+2.19)  \\
\hhline{~----}
& Extra layer & C & 93.8 (-0.3) & 92.2 (-0.48) \\
\hline
\multirow{2}{*}{\parbox{1.5cm}{\centering ImageNet fruit vs. vege}} & Baseline CNN & F & 95.4 (+3.6) & 93.15 (+3.5)  \\
\hhline{~----}
& Extra layer & C & 91.7 (-0.1) & 89.67 (+0.02) \\
\hline
\end{tabular}
\end{center}

\vspace{-15pt}
\end{table}

\subsection{Generalization}\label{sec:generalization}

As shown in both Table \ref{table:prior_new} and Figure \ref{fig:training_curve}, training with fine-grain labels (vs. coarse-grain) achieves higher testing accuracy. This may partially be due to better network optimization, because under ImageNet subsets, fine-grain training improves both training and testing accuracy. However, in the cases of CIFAR-10, CIFAR-100 animals and CIFAR-100, even for the same training accuracy, testing accuracy for fine-grain trained CNNs is still higher than coarse-grain training. This indicates that fine-grain training delivers higher generalization capability. 

Our intuition is that with fine-grain labels, the CNN is able to learn more features than training with coarse-grain labels. For example, suppose that all cat images in the training set have whiskers, while none of the dogs has whiskers. Then, as long as the network trained with coarse-grain labels learns this feature, it can produce $100\%$ training accuracy with no need to learn any other features. This is a well known phenomenon in weakly-supervised learning, in which the network only learns the most discriminative features~\cite{bilen2016weakly}. Then, in the testing set, if a cat image does not include whiskers, the network will make an incorrect prediction. However, with fine-grain labels, the network needs to learn more features (\textit{e.g.}, ears, tails, etc.) to distinguish among different breeds of dogs and cats. These extra features learned through fine-grain labeling may help the network's performance on coarse-grain class classification on the testing set, \textit{e.g.}, it now can tell if it is a cat through ears, tails, etc, even though it does not have whiskers.

\begin{figure}[ht]
    \centering
    \subfloat[][t-SNE visualization of two coarse-grain classes in CIFAR-10 trained with two coarse-grain labels]{\includegraphics[width=.45\linewidth]{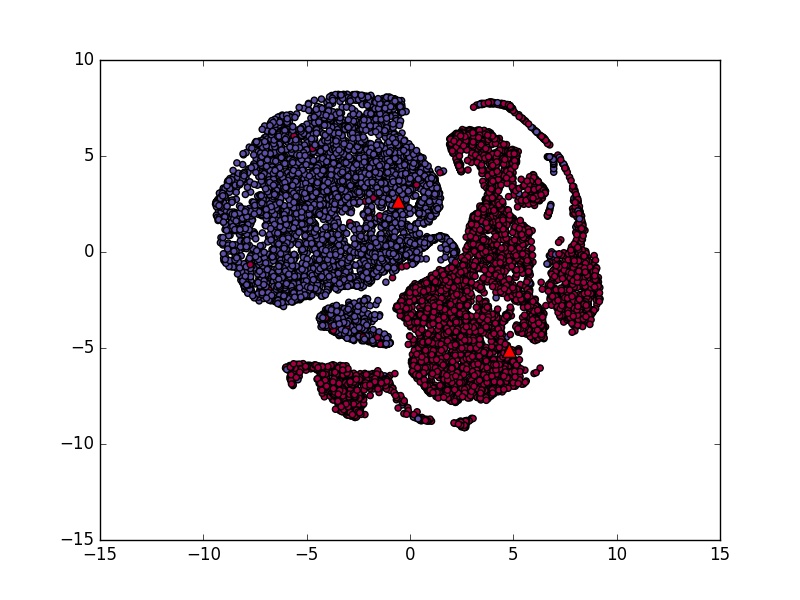}\label{fig:tsne_cifar10_coarse}}
    \hspace{8pt}
    \subfloat[][t-SNE visualization of two coarse-grain classes in CIFAR-10 trained with ten fine-grain labels]{\includegraphics[width=.45\linewidth]{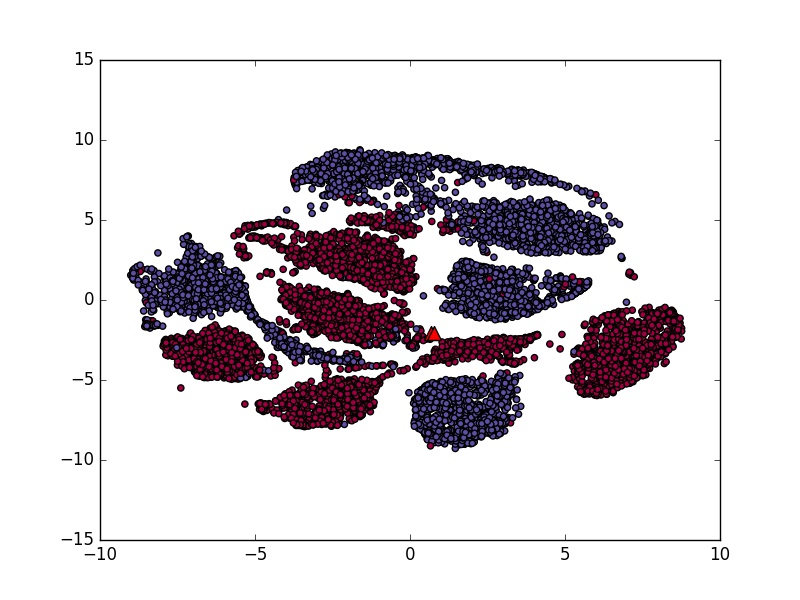}\label{fig:tsne_cifar10_fine}}
    \caption{t-SNE visualization of CIFAR-10 test set trained with coarse-grain labels vs. fine-grain labels. Data points shown in the same color belong to the same coarse-grain class.}
    \label{fig:ttsne_cifar10}
\end{figure}

Figure \ref{fig:ttsne_cifar10} shows the t-distributed Stochastic Neighbor Embedding (t-SNE) visualization \cite{maaten2008visualizing} of all CIFAR-10 testing images with coarse-grain (a) and fine-grain training (b). 
\vspace{-5pt}

Image features used for t-SNE visualization are the outputs of the second-to-last fully-connected layer, which is a technique commonly used to extract compact semantic representation of the raw input images~\cite{maaten2008visualizing}. These feature vectors are then transformed by the t-SNE technique~\cite{maaten2008visualizing} to a two-dimensional space for visualization. All data points are colored according to their ground-truth coarse-grain labels. We also show the position of means of each coarse-grain class as red triangles in the figures. We can see that, for both coarse-grain training, Figure \ref{fig:tsne_cifar10_coarse}, and fine-grain training, Figure \ref{fig:tsne_cifar10_fine}, there is a noticeable margin between coarse-grain classes, and a decision boundary can be drawn to separate them. However, the network has to learn extra features to further separate the fine-grain classes within each coarse-grain class when trained with fine-grain labels (as shown in Figure \ref{fig:tsne_cifar10_fine}), while when trained with only coarse-grain labels, the data points are merged together as there is no need to separate them (as being visualized in Figure \ref{fig:tsne_cifar10_coarse}).


An orthogonal method used for enhancing the variety of learned features and thereby increasing generalization ability is dropout~\cite{srivastava2014dropout}. Dropout randomly drops some of the features to encourage CNN to learn more various features. A possible question arises: by adding dropout to the network, will coarse-grain training reach the same testing accuracy as fine-grain training? 

In cases of CIFAR-100 and CIFAR-100 animals, the original network already has dropout layers within each residual block with the optimal dropout rate 0.3 determined by experiments~\cite{zagoruyko2016wide}. However, as shown in Table \ref{table:prior_new}, fine-grain training still outperforms coarse-grain training, under optimal dropout rate, which indicates that fine-grain training delivers benefits that dropout alone may not.
We further conduct experiments on CIFAR-10 and ImageNet subsets, by adding a dropout layer between the global pooling layer and the fully-connected layer. The dropout rate is set as 0.3 in our following experiments.  

Table \ref{table:add_dropout} shows the experimental results of using the dropout technique. We observe that adding the dropout layer provides limited improvement in testing accuracy for coarse-grain training, and dropout for coarse-grain training still generates a noticeable margin when compared to the fine-grain training with or without dropout. This indicates that fine-grain labels can further improve CNN learning beyond what the traditional dropout technique can do. Actually, since fine-grain training and dropout are two orthogonal techniques, one can use both to further improve CNN performance. For example, in ImageNet subsets dog vs.cat and fruit vs. vegetable, combining the two techniques can able to push the testing accuracy to 95\% and 93.86\% from 92.68\% and 89.65\%, respectively.

\begin{table}[ht]
\begin{center}
\caption{Experiments on increasing CNN dropout rate. Values in ``Dropout" column indicates dropout rates used. In ``Train Label" column: ``F" and ``C" indicate fine-grain and coarse-grain labels, respectively.}
\vspace{3pt}
\label{table:add_dropout}
\begin{tabular}{|C{1.6cm}|C{1.6cm}|C{0.9cm}|C{1.2cm}|C{1.2cm}|}
\hline
Dataset & Dropout & Train Label & Training accuracy (\%) & Testing accuracy (\%) \\
\hline
\multirow{4}{*}{CIFAR-10} & 0.3 & F & 100.0 & 99.10  \\
\hhline{~----}
& 0.3 & C & 99.9 & 98.48  \\
\hhline{~----}
& 0 & F & 100.0 & 99.20  \\
\hhline{~----}
& 0 & C & 99.9 & 98.42  \\
\hline
\multirow{4}{*}{\parbox{1.5cm}{\centering ImageNet dog vs. cat}} & 0.3 & F & 94.8 & 95.00 \\
\hhline{~----}
& 0.3 & C & 94.3 & 92.80 \\
\hhline{~----}
& 0 & F & 95.3 & 94.87  \\
\hhline{~----}
& 0 & C & 94.1 & 92.68  \\
\hline
\multirow{4}{*}{\parbox{1.5cm}{\centering ImageNet fruit vs. vege}} & 0.3 & F & 95.0 & 93.86  \\
\hhline{~----}
& 0.3 & C & 91.7 & 89.93 \\
\hhline{~----}
& 0 & F & 95.4 & 93.15  \\
\hhline{~----}
& 0 & C & 91.8 & 89.65  \\
\hline
\end{tabular}
\end{center}

\vspace{-15pt}
\end{table}

\section{Characterizing the effectiveness of fine-grain labels: Average Confusion Ratio} \label{sec:acr}

\begin{figure}[t]
    \centering
    \includegraphics[width=3.5in]{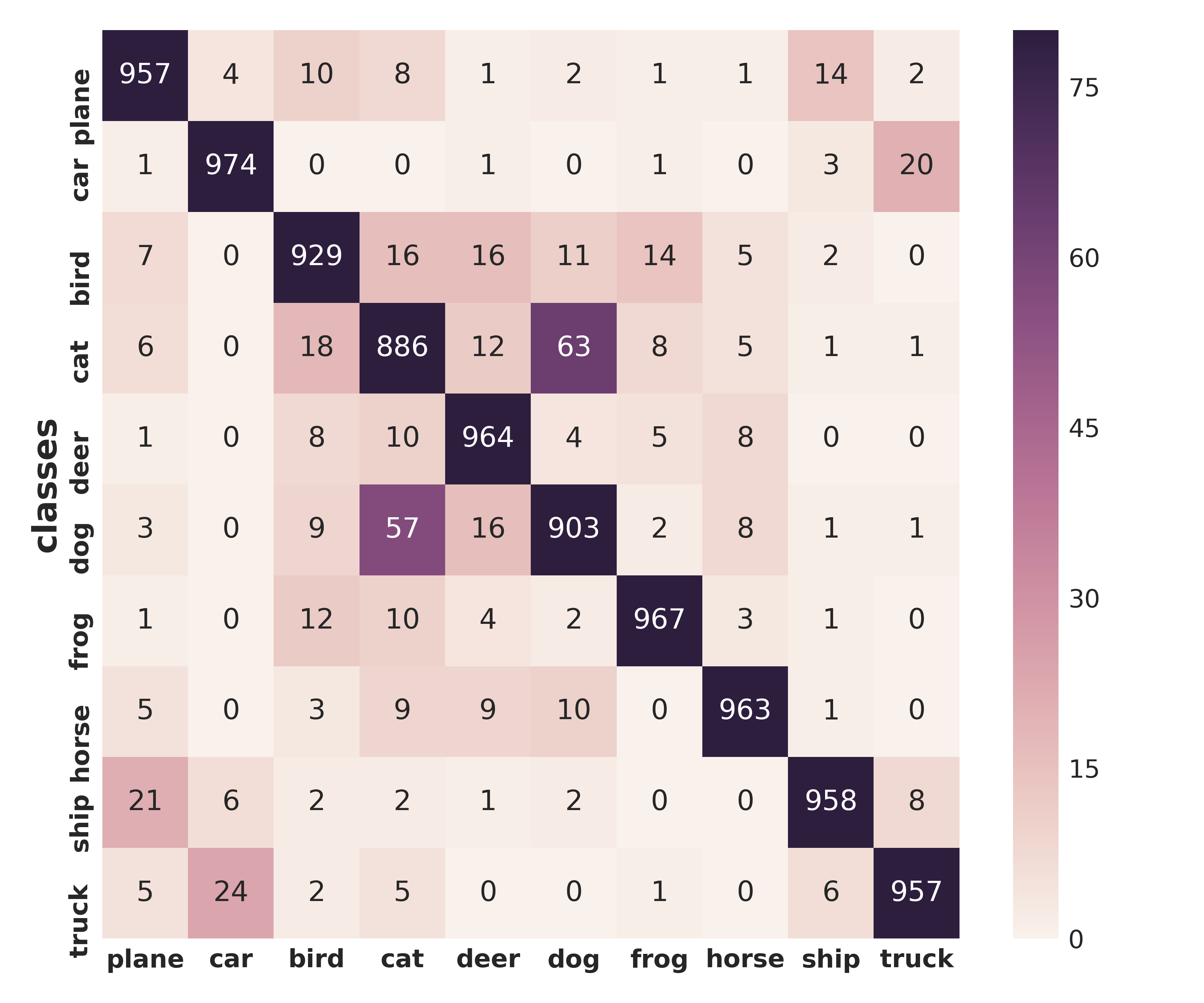}
    \caption{Confusion matrix for ten classes of CIFAR-10 dataset.}
    \label{fig:conf_matrix}
\end{figure}

Fine-grain labels can improve both CNN optimization and generalization as shown by the experiments in the previous sections. However, we also note the varying benefit from fine-grain label usage under different datasets: fine-grain training sometimes improves testing accuracy by a considerably large margin, \textit{e.g.}, 3.5\% improvement in ImageNet fruit vs. vegetable, while sometimes the improvement is rather limited, \textit{e.g.}, 0.01\% improvement in CIFAR-100. Similar to the inter- and intra-cluster variance used in unsupervised clustering algorithms, \textit{e.g.}, $k$-means~\cite{han2011data}, the benefit from fine-grain training may come from the relative difficulty of distinguishing between coarse-grain classes (inter-class confusion) vs. fine-grain classes (intra-class confusion). To quantify this, we propose the \textit{Average Confusion Ratio} (ACR) metric to characterize the disparity within the coarse-grain and fine-grain classes, respectively, by using the confusion matrix shown in Fig \ref{fig:conf_matrix}. We denote the \textit{confusion matrix} as $\mathbb{C}$, where $\mathbb{C}_{i,j}$ indicates the number of occurrences of confusing class $i$ with class $j$ and it can be obtained via counting those occurrences through the test dataset~\cite{sokolova2009systematic}. From the confusion matrix $\mathbb{C}$ for the fine-grain classes as in Figure \ref{fig:conf_matrix}, we can compute the ACR: 
\begin{equation}
    ACR = \frac{\sum_{(i,j)\in \overline{\mathbb{A}}} \mathbb{C}_{i,j} / |\overline{\mathbb{A}}|} {\sum_{(i,j)\in \mathbb{A}} \mathbb{C}_{i,j} / |\mathbb{A}|},
\end{equation}
where $\mathbb{A} = \{ (i,j)| \mathbb{B}_{i,j} = 1\}$, $\overline{\mathbb{A}}=\{ (i,j)| \mathbb{B}_{i,j} = 0\}$, and $\mathbb{B}$ is an indicator matrix with $\mathbb{B}_{i,j}$ indicating whether class $i$ and $j$ belong to the same coarse-grain class. Intuitively, ACR is the average inter-class confusion divided by average intra-class confusion, where inter- and intra-classes are considered from the perspective of coarse-grain classes. 

ACR is correlated to the improvement produced by fine-grain training. We define the improvement from fine-grain training as the difference between the testing accuracy of a CNN trained with fine-grain labels and the testing accuracy of a CNN trained with coarse-grain labels, \textit{i.e.}, $\Delta A^{test} = A_{FC}^{test} - A_{CC}^{test}$. Lower ACR means lower relative confusion across coarse-grain classes and hence higher distance between coarse-grain classes. This is a similar concept to high inter-cluster distance in clustering algorithms~\cite{fradkin2008clustering}\cite{jain1999data}, and those clusters are less prone to be mixed or confused. As a result, coarse-grain classes in this case are relatively easier to be separated even without the help of fine-grain labels, which leads to a low $\Delta A^{test}$ value, and vice versa.

\begin{figure}[ht]
    \centering
    \includegraphics[width=0.8\linewidth]{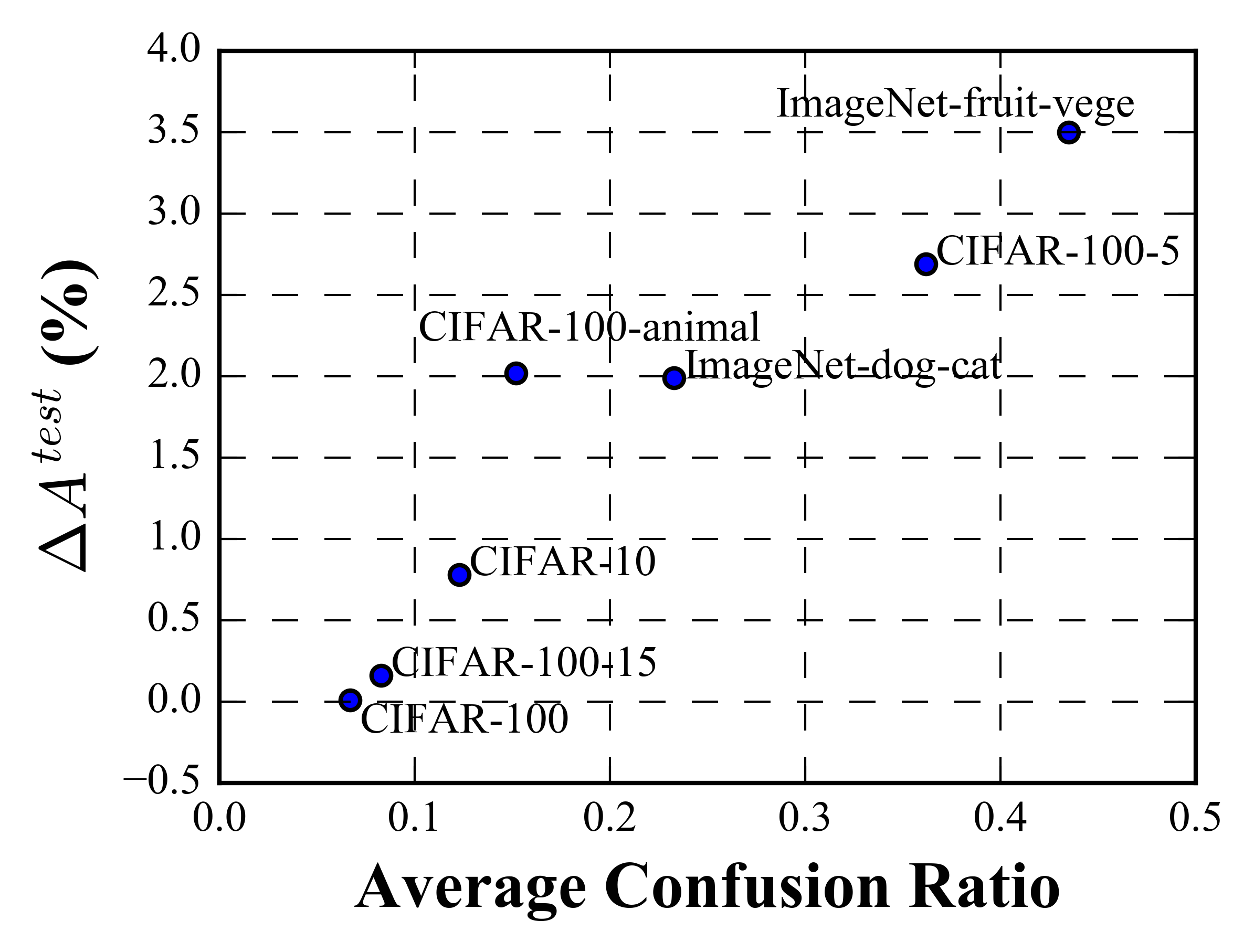}
    \caption{$\Delta A^{test}$ vs. Average Confusion Ratio across different datasets. }
    \label{fig:ACR_datasets_added}
    
    \vspace{-15pt}
\end{figure}

To demonstrate how ACR can be an indicator of how much improvement fine-grain labels deliver in different datasets, we compute the ACR metric for all datasets in Table~\ref{table:prior_new} and plot the relationship between ACR and $\Delta A^{test}$ in Figure~\ref{fig:ACR_datasets_added}. In general the data points in Figure~\ref{fig:ACR_datasets_added} show higher ACR leading to higher $\Delta A^{test}$, as expected. Other than the five datasets used throughout the paper, we also introduce two extra datasets: CIFAR-100-5 and CIFAR-100-15 with 5 and 15 coarse-grain classes, respectively. In the next section, we will detail these two datasets and the corresponding ACR metric under different settings of coarse- and fine-grain classes.

\begin{table*}[ht]
\begin{center}
\caption{Testing accuracy, trained with coarse-grain vs. fine-grain labels, of customized coarse-grain classes of CIFAR-10 dataset. Zero and one indicates which coarse-grain class each fine-grain class belongs to.}
\vspace{3pt}
\label{table:random_coarse}
\begin{tabular}{|c|c|c|c|c|c|c|c|c|c|c|c|c|c|c|}
\hline
\multirow{2}{*}{ID} & \multirow{2}{*}{Ratio} & \multicolumn{10}{c|}{Classes} & \multirow{2}{*}{$A_{CC}^{test}$ (\%)} & \multirow{2}{*}{$A_{FC}^{test}$ (\%)} & \multirow{2}{*}{$\Delta A^{test}$ (\%)}\\
\hhline{~~----------~~}
& & plane & car & bird & cat & deer & dog & frog & horse & ship & truck & & \\
\hline
(1) & \multirow{5}{*}{6:4} & 0 & 0 & 1 & 1 & 1 & 1 & 1 & 1 & 0 & 0 & 98.42 & 99.20 & +0.78\\
\hhline{-~-------------}
(2) & & 1 & 0 & 1 & 1 & 1 & 1 & 0 & 1 & 0 & 0 & 97.68 & 98.64 & +0.96\\
\hhline{-~-------------}
(3) & & 0 & 0 & 0 & 1 & 1 & 1 & 1 & 1 & 1 & 0 & 96.95 & 98.02 & +1.07\\
\hhline{-~-------------}
(4) & & 0 & 0 & 1 & 0 & 1 & 0 & 1 & 1 & 1 & 1 & 95.26 & 97.20 & +1.94\\
\hhline{-~-------------}
(5) & & 0 & 0 & 1 & 0 & 0 & 1 & 1 & 1 & 1 & 1 & 93.44 & 96.22 & +2.78\\
\hline
(6) & \multirow{5}{*}{5:5} & 0 & 0 & 0 & 1 & 1 & 1 & 1 & 1 & 0 & 0 & 97.60 & 98.51 & +0.91\\
\hhline{-~-------------}
(7) & & 0 & 0 & 1 & 0 & 1 & 1 & 1 & 1 & 0 & 0 & 95.90 & 97.59 & +1.69\\
\hhline{-~-------------}
(8) & & 0 & 1 & 0 & 1 & 0 & 1 & 1 & 1 & 0 & 0 & 96.17 & 97.54 & +1.37\\
\hhline{-~-------------}
(9) & & 0 & 0 & 1 & 0 & 0 & 1 & 0 & 1 & 1 & 1 & 94.15 & 96.28 & +2.13\\
\hhline{-~-------------}
(10) & & 1 & 0 & 0 & 0 & 0 & 1 & 1 & 1 & 0 & 1 & 94.19 & 96.16 & +1.97\\
\hline
\end{tabular}
\end{center}

\vspace{-15pt}
\end{table*}

\begin{table}[ht]
\begin{center}
\caption{Testing accuracy trained with noisy fine-grain labels of CIFAR-10 dataset.}
\vspace{3pt}
\label{table:random_fine}
\begin{tabular}{|c|c|c|}
\hline
Randomness factor & $A_{FC}^{test}$ (\%) & $\Delta A^{test}$ (\%) \\
\hline
0 & 99.20 & 0.78 \\
\hline 
0.01 & 98.94 & 0.52 \\
\hline
0.03 & 98.55 & 0.13 \\
\hline
0.1 & 98.12 & -0.30 \\
\hline
0.3 & 97.72 & -0.70 \\
\hline
\end{tabular}
\end{center}

\vspace{-15pt}
\end{table}

\section{Discussion} \label{sec:discuss}
In this section, we further explore several scenarios in which the setting of coarse-grain and fine-grain labels change. More specifically, coarse-grain classes may vary due to the requirement of the application and the fine-grain labels may be noisy if it is generated via automatic unsupervised clustering algorithms. Again, we show that ACR is able to capture these characteristics and correctly reflect the effect of fine-grain training. We also investigate how increasing the number of coarse-grain classes impacts the improvement from using fine-grain labels, \textit{i.e.}, $\Delta A^{test}$.

In the following experiments, we use CIFAR-10 as an example to show how ACR can be used to characterize the effectiveness of fine-grain labels via the relationship between ACR and $\Delta A^{test}$ under different settings of coarse-grain and fine-grain labels. We use CIFAR-100 for the experiments on varying number of coarse-grain classes as it provides as many as 20 coarse-grain classes.

\subsection{Customized Coarse-grain Classes}
As mentioned, coarse-grain classes are the classification target, and as a result, the
definition of coarse-grain classes is application dependent. For example, given an animal dataset, a task can be identifying cat vs. dog. vs horse, while another task can be separating standing animals from sitting and/or lying animals. Because of the diversity of applications, this mapping from fine-grain classes to coarse-grain classes can be drastically different. In this section, we conduct experiments to see how these customized coarse-grain classes affect the effectiveness of fine-grain labels and use ACR to characterize it.

A natural partition of CIFAR-10 dataset is the ``animal" coarse-grain class vs. the ``vehicle" coarse-grain class, where ``animal" has six fine-grain classes and ``vehicle" has four as depicted in Table \ref{table:class_setting}. To simulate various applications, we keep the 6:4 ratio of the two coarse-grain classes and randomly switch their fine-grain classes to create new coarse-grain classes. Rows (1) through (5) in Table \ref{table:random_coarse} show five experiments with different coarse-grain class definitions. We use two coarse-grain classes in this case (denoted by 0 and 1), and values in the table indicate which coarse-grain class (0 vs. 1) each fine-grain class (plane, car, etc.) belongs to. The last three columns of Table \ref{table:random_coarse} give the testing accuracy of the CNN trained with coarse-grain and fine-grain labels, respectively as well as the relative improvement of fine-grain training. We observe that fine-grain training achieves up to 2.78\% improvement and always outperforms coarse-grain training under various customized coarse-grain classes.

We further experiment with balanced coarse-grain classes. In the previous experiments, we have a 6:4 ratio for the number of fine-grain classes within each coarse-grain class. Now, we balance it to a 5:5 ratio, and similarly, we randomly switch fine-grain classes across the two coarse-grain classes. Rows (6) through (10) in Table \ref{table:random_coarse} show five experiments with different coarse-grain class definitions and a 5:5 ratio. Again, we can see that fine-grain training always produces higher testing accuracy than coarse-grain training.

As discussed before, higher ACR leads to higher $\Delta A^{test}$ and vice versa.
We compute the ACR metric of all ten experiments and plot the relationship between ACR and $\Delta A^{test}$ in Figure \ref{fig:cifar10_custom_coarse}. Numbers on the data points are experiment IDs. We can see the trend of increasing benefit from fine-grain labeling, \textit{i.e.}, increasing $\Delta A^{test}$, when ACR gets larger. This demonstrate that ACR is a good indicator of how effective fine-grain labels are.

\subsection{Noisy Fine-grain Classes} \label{sec:random_fine_class}
By using fine-grain labels, we are able to improve CNN performance. To obtain fine-grain labels, we can either ask human to label the images, or by automatically clustering every coarse-grain class into multiple fine-grain classes. The first approach is human-labor intensive but it is usually defined as the ground-truth, while the second approach is relatively cheap, but error-prone. In this part, we investigate how a noisy fine-grain label, \textit{e.g.}, generated from a coarse-grain class by using unsupervised clustering methods, may affect effectiveness of training with fine-grain labels.

To this end, we keep the coarse-grain labels fixed and randomly change the fine-grain labels within each coarse-grain class to simulate the effect of noisy labeling. We tune the probability of randomizing the fine-grain labels, \textit{i.e.}, randomness factor, to control the amount of noise in the experiments. Table \ref{table:random_fine} shows the results under different randomness factors for CIFAR-10 dataset. We can see that with increased randomness factor, both $A_{FC}^{test}$ and the improvement brought by fine-grain training, $\Delta A^{test} = A_{FC}^{test} - A_{CC}^{test}$, keep dropping. This means that training with highly incorrect fine-grain labels may actually hurt CNN performance. Therefore, how to automatically cluster each coarse-grain class into less-noisy fine-grain classes is an important direction to explore. We leave it for future work.

\begin{figure}[ht]
    \centering
    \subfloat[][]{\includegraphics[width=.78\linewidth]{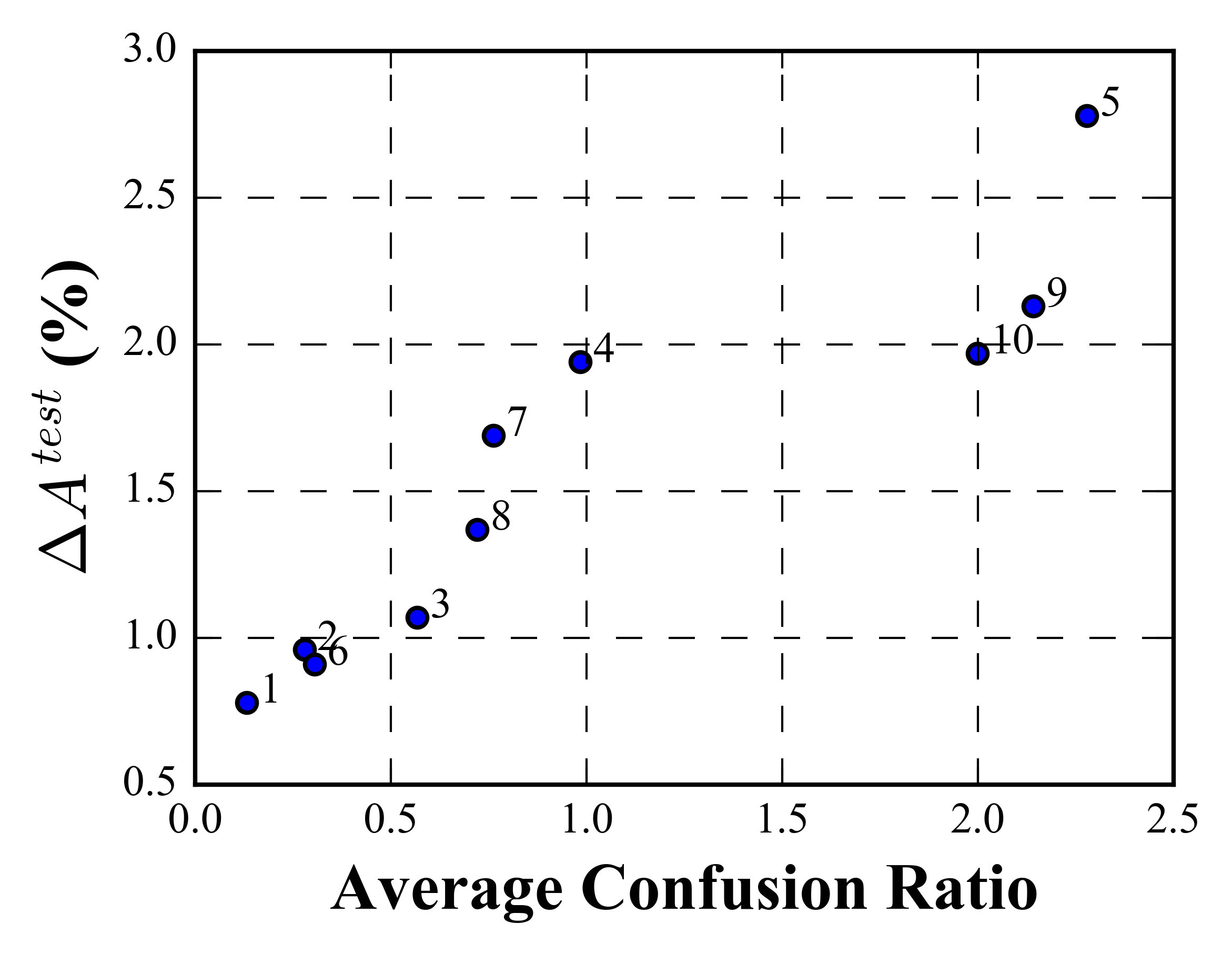}\label{fig:cifar10_custom_coarse}}
    \vspace{-3pt}
    
    \subfloat[][]{\includegraphics[width=.78\linewidth]{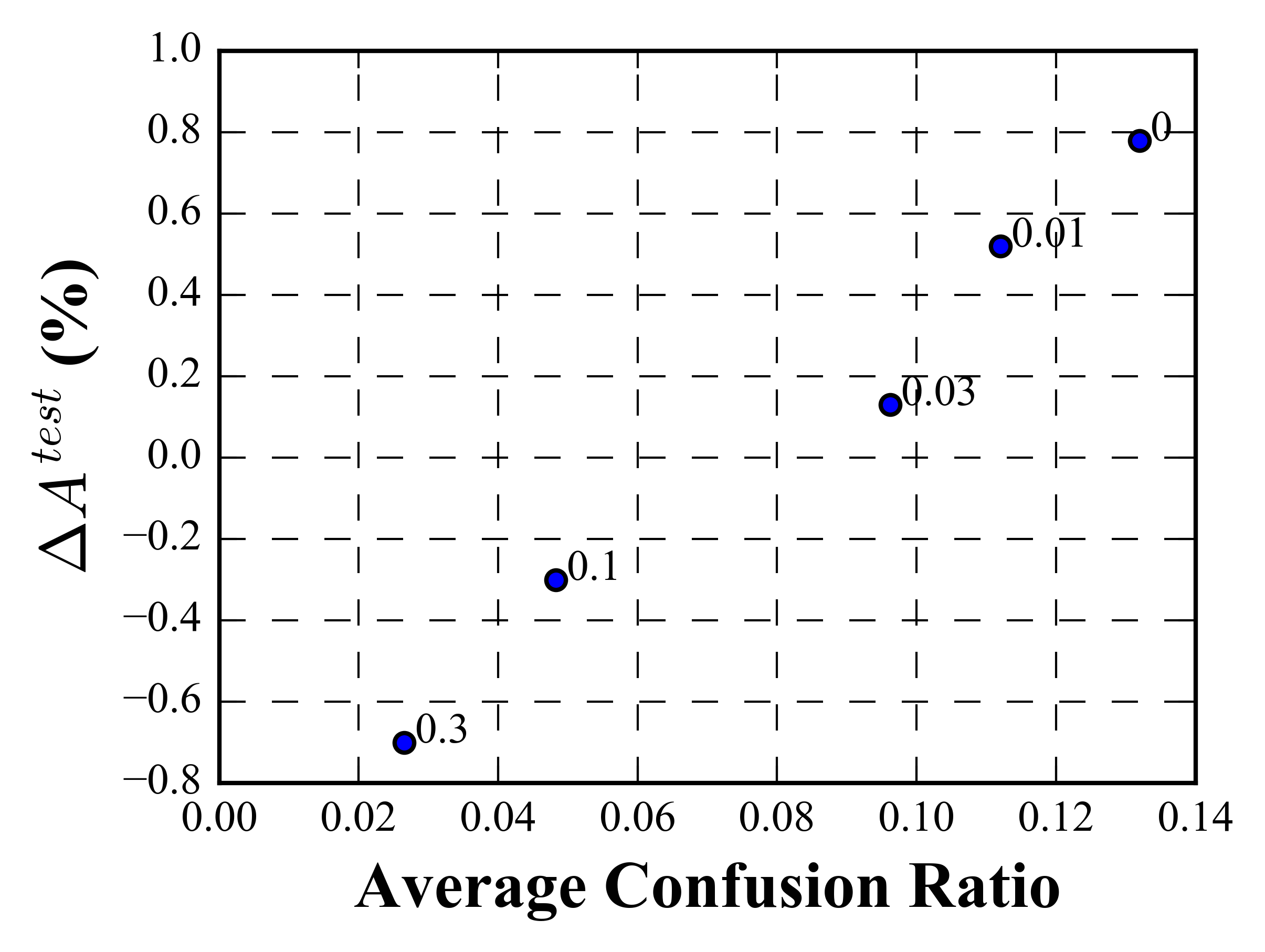}\label{fig:cifar10_random_fine}}
    \caption{(a): $\Delta A^{test}$ vs. Average Confusion Ratio computed from experiments of customized coarse-grain classes in Table \ref{table:random_coarse} . Number next to the data points is experiment ID. (b): $\Delta A^{test}$ vs. Average Confusion Ratio computed from experiments of noisy fine-grain classes in Table \ref{table:random_fine}. Number next to the data points is randomness factor.}
    \label{fig:cifar10_ACR_random}
    
    \vspace{-15pt}
\end{figure}

Again, we compute their ACR values and plot $\Delta A^{test}$ vs. ACR in Figure \ref{fig:cifar10_random_fine}. Numbers on the data points are randomness factors. With decreased randomness factor, confusion between fine-grain classes becomes less and ACR value increases. As expected, increased ACR value leads to increased $\Delta A^{test}$ as we can see in the figure. 
 
\begin{table*}[!htb]
\begin{center}
\caption{Testing accuracy, trained with coarse-grain vs. fine-grain labels, when varying number of coarse-grain classes in CIFAR-100 dataset. The coarse-grain class index follows the same order as in Table \ref{table:class_setting}. The values inside the parenthesis in column $A_{FC}^{test}$ is $\Delta A^{test}$, the calculated improvement of fine-grain training over coarse-grain training.}
\vspace{3pt}
\label{table:coarse_number}
\begin{tabular}{|c|c|c|c|c|c|c|c|c|c|c|c|c|c|c|c|c|c|c|c|c|c|c|}
\hline
\multicolumn{20}{|c|}{Coarse-grain class index} & \multirow{2}{*}{Total} & \multirow{2}{*}{$A_{CC}^{test}$ (\%)} & \multirow{2}{*}{$A_{FC}^{test}$ (\%)} \\
\hhline{--------------------}
0 & 1 & 2 & 3 & 4 & 5 & 6 & 7 & 8 & 9 & 10 & 11 & 12 & 13 & 14 & 15 & 16 & 17 & 18 & 19 & & & \\
\hline
\checkmark &  &  &  &  &  &  &  & \checkmark &  &  & \checkmark & \checkmark &  &  &  & \checkmark &  & & &  5 & 80.53 & 83.22 (+2.69)  \\
\hline
\checkmark & \checkmark &  &  &  &  &  & \checkmark & \checkmark &  &  & \checkmark & \checkmark & \checkmark & \checkmark & \checkmark & \checkmark &  &  & & 10 & 81.42 & 83.44 (+2.02) \\
\hline
\checkmark & \checkmark & \checkmark &  & \checkmark &  &  & \checkmark & \checkmark &  &  & \checkmark & \checkmark & \checkmark & \checkmark & \checkmark & \checkmark & \checkmark & \checkmark & \checkmark & 15 & 85.14 & 85.30 (+0.16) \\
\hline
\checkmark & \checkmark & \checkmark & \checkmark & \checkmark & \checkmark & \checkmark & \checkmark & \checkmark & \checkmark & \checkmark & \checkmark & \checkmark & \checkmark & \checkmark & \checkmark & \checkmark & \checkmark & \checkmark & \checkmark & 20 & 85.04 & 85.05 (+0.01) \\
\hline

\end{tabular}
\end{center}

\vspace{-15pt}
\end{table*}

\subsection{Varying number of coarse-grain classes}\label{sec:vary_coarse}
We further investigate how the number of coarse-grain classes affects the effectiveness of fine-grain labels. As we have discussed in Section IV.B, with fine-grain labels, the network is encouraged to learn more features than it needs when trained with only coarse-grain labels, and these extra features help in network generalization, \textit{i.e.}, improving the testing accuracy. We conjecture that, to achieve high testing accuracy, a certain number of features needs to be learned by the network. Fine-grain labels help learn more features, however, with more coarse-grain classes, more features will be learned from only coarse-grain labels and hence it may be sufficient for classifying the test set, even without fine-grain labels. In other words, fine-grain labels bring diminishing returns when the number of coarse-grain classes increases. 

To verify this, we experiment by varying the number of coarse-grain classes in the CIFAR-100 dataset and the results are shown in Table \ref{table:coarse_number}. We can see that with increasing number of coarse-grain classes, \textit{i.e.}, from 5, 10, 15 to 20, the benefit from fine-grain training, \textit{i.e.}, $\Delta A^{test}$, decreases, which is consistent with our expectation. We also compute their ACR values and show the relationship with $\Delta A^{test}$ in Figure \ref{fig:ACR_datasets_added}. In the case of CIFAR-100 dataset, when the number of coarse-grain classes goes beyond 15, the improvement brought by fine-grain labeling is negligible. However, this threshold is application and dataset dependent and should be determined by experiments in a case-by-case manner.

\section{Conclusion} \label{sec:conclusion}
In this paper, we investigate the intriguing problem of how label granularity impacts CNN-based image classification. Our extensive experimentation shows that using fine-grain labels, rather than the target coarse-grain labels, can lead to higher accuracy and training data efficiency by improving both network optimization and generalization. Our results further suggest two practical applications: (i) with sufficient human resources, one can improve CNN accuracy by re-labeling the dataset with fine-grain labels, and (ii) with limited human resources, to improve CNN performance, rather than collecting more training data, one may instead collect fine-grain labels for the existing data. Furthermore, we propose a metric called \textit{Average Confusion Ratio} (ACR) to quantify the accuracy gain from fine-grain labels, and demonstrate its effectiveness through experiments on various datasets and label settings.

{\small
\bibliographystyle{IEEEtran}
\bibliography{egbib}
}

\end{document}